\PassOptionsToPackage{dvipsnames,table}{xcolor}
\documentclass[10pt,twocolumn,letterpaper]{article}

\usepackage[pagenumbers]{cvpr} 
\usepackage[accsupp]{axessibility}

\usepackage{ifxetex}
\ifxetex
  \usepackage{ctex}
\fi

\definecolor{phcolor}{RGB}{188, 188, 188}
\usepackage{lipsum}
\usepackage{fontawesome5}
\usepackage{balance}

\usepackage{multirow}
\usepackage{hhline}
\usepackage[listings]{tcolorbox}
\usepackage{enumitem}
\usepackage{pifont}
\usepackage{dsfont}
\usepackage{balance}
\usepackage{mathrsfs}
\usepackage{diagbox}

\definecolor{Gray}{gray}{0.9}
\definecolor{White}{gray}{1}
\definecolor{DGray}{gray}{0.8}
\definecolor{DDDDGray}{gray}{0.3}
\definecolor{WhiteGray}{rgb}{0.9, 0.9, 0.9}
\definecolor{citecolor}{HTML}{0071bc}
\definecolor{darkred}{rgb}{0.6, 0.1, 0.05}
\definecolor{DeltaColor}{rgb}{0.039,0.73,0.71}
\definecolor{SigmaColor}{rgb}{0.98,0.45,0.0}
\definecolor{AlphaColor}{rgb}{0,0,0.8}
\definecolor{BetaColor}{rgb}{0.8,0,0.8}
\definecolor{GammaColor}{rgb}{0.514,0.34,0.224}
\definecolor{EpsilonColor}{rgb}{0.353,0.725,0.906}
\definecolor{PurpleColor}{HTML}{8B008B}
\definecolor{BadColor}{HTML}{C0392B}
\definecolor{OrangeColor}{rgb}{0.914,0.541,0.0.141}
\definecolor{GreenColor}{rgb}{0.137,0.573,0.565}
\definecolor{RedColor}{rgb}{0.949,0.275, 0.224}
\definecolor{LightCyan}{rgb}{0.88,1,1}
\definecolor{Gray}{gray}{0.85}

\definecolor{bestcolor}{rgb}{1, 0.5, 0.25}
\definecolor{secondbestcolor}{rgb}{1, 0.8, 0.5}

\newcommand{\del}[1]{}

\DeclareMathAlphabet\mathbfcal{OMS}{cmsy}{b}{n}

\newcommand\supp{Appx\xspace}

\newcommand{\qheading}[1]{\noindent\mbox{\textbf{#1}\;}}

\newcommand{\greencheck}{{\color{GreenColor} \large \ding{51}}\xspace}
\newcommand{\bluecheck}{{\color{blue} \checkmark}\xspace}

\newlength\savewidth

\newcommand{\projectURL}{\url{https://maniptrans.github.io}}

\newcommand{\method}{\textbf{\textsc{ManipTrans}}\xspace}
\newcommand{\dataset}{\textbf{\textsc{DexManipNet}}\xspace}

\newcommand{\mbold}[1]{\boldsymbol{#1}}
\newcommand{\obj}{\mbold{o}}
\newcommand{\aobj}{o^\text{A}}
\newcommand{\caobj}{{\hat{o^\text{A}}}}
\newcommand{\cobj}{\mbold{\hat{o}}}
\newcommand{\hand}{\mbold{h}}
\newcommand{\dexhand}{\mbold{d}}

\newcommand{\Traj}{\mbold{\mathcal{T}}}
\newcommand{\traj}{\mbold{\tau}}
\newcommand{\imitator}{\mbold{\mathcal{I}}}
\newcommand{\residual}{\mbold{\mathcal{R}}}
\newcommand{\Reward}{\mbold{\mathsf{R}}}

\definecolor{cvprblue}{rgb}{0.21,0.49,0.74}
\usepackage[pagebackref,breaklinks,colorlinks,allcolors=cvprblue]{hyperref}

\def\paperID{1611} 
\def\confName{CVPR}
\def\confYear{2025}

\title{\texorpdfstring{\method: Efficient Dexterous Bimanual Manipulation Transfer \\via Residual Learning}{\method: Efficient Dexterous Bimanual Manipulation Transfer via \del{Incremental}Residual Learning}}

\author{
    {
        Kailin Li\textsuperscript{1}\;
        Puhao Li\textsuperscript{1,2}\;
        Tengyu Liu\textsuperscript{1}\;
        Yuyang Li\textsuperscript{1,3}\;
        Siyuan Huang\textsuperscript{1}\; 
    } \\
    { 
      \small
      $^{1}$State Key Laboratory of General Artificial Intelligence, BIGAI
    } \\
    {
      \small
      $^{2}$Department of Automation, Tsinghua University
      \quad\quad
      $^{3}$Institute for Artificial Intelligence, Peking University
    } \\
    {
        \small
        \projectURL
    }
}

\begin{document}

\newcommand{\teaserCaption}{
\textbf{\method for Bimanual Dexterous Manipulations.}
Retargeting methods often struggle with transferring MoCap data to physically plausible motions, while our \method efficiently produces task-compliant, physically accurate motions. It also generalizes across embodiments like Inspire hands~\cite{inspirehandurl}, Shadow hands~\cite{shadowhandurl}, and articulated MANO hands~\cite{MANO:SIGGRAPHASIA:2017, christen2022d}.\vspace{-0.8em}}

\twocolumn[{
    \renewcommand\twocolumn[1][]{#1}
    \maketitle
    \centering
    \vspace{-1.8em}
    \begin{minipage}{1.00\textwidth}
        \centering
        \colorbox{white}{\includegraphics[trim=000mm 000mm 000mm 000mm, clip=False, width=\linewidth]{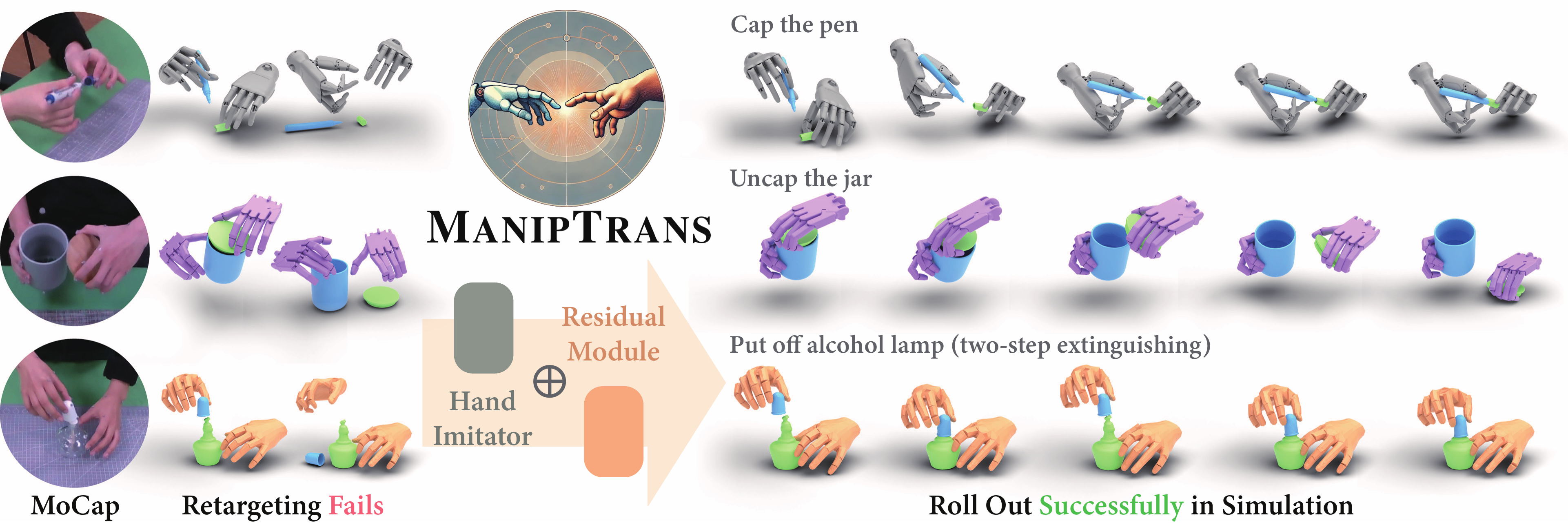}}
    \end{minipage}
    \captionsetup{type=figure}
    \captionof{figure}{\teaserCaption}
    \label{fig:teaser}
    \vspace*{+1.8 em}
}]

\begin{abstract}

    \vspace{-1.2em}

    Human hands play a central role in interacting, motivating increasing research in dexterous robotic manipulation. Data-driven embodied AI algorithms demand precise, large-scale, human-like manipulation sequences, which are challenging to obtain with conventional reinforcement learning or real-world teleoperation. To address this, we introduce \textnormal{\method}, a novel two-stage method for efficiently transferring human bimanual skills to dexterous robotic hands in simulation. \textnormal{\method} first pre-trains a generalist trajectory imitator to mimic hand motion, then fine-tunes a specific residual module under interaction constraints, enabling efficient learning and accurate execution of complex bimanual tasks.
    Experiments show that \textnormal{\method} surpasses state-of-the-art methods in success rate, fidelity, and efficiency. Leveraging \textnormal{\method}, we transfer multiple hand-object datasets to robotic hands, creating \textnormal{\dataset}, a large-scale dataset featuring previously unexplored tasks like pen capping and bottle unscrewing. \textnormal{\dataset} comprises 3.3K episodes of robotic manipulation and is easily extensible, facilitating further policy training for dexterous hands and enabling real-world deployments.

    \vspace{-1.8em}

\end{abstract}    
\section{Introduction}
\label{sec:intro}

Embodied AI (EAI) has advanced rapidly in recent years, with increasing efforts to enable AI-driven embodiments to interact with physical or virtual environments. Just as human hands are pivotal for interaction, much research in EAI focuses on dexterous robotic hand manipulation~\cite{agarwal2023dexterous, chen2023sequential, wang2024dexcap, lu2024garmentlab, mao2024dexskills, yuan2024cross, chen2022system, chen2024vegetable, chen2023visual, liconti2024leveraging, li2024okami, liu2024masked, she2024learning, chenobject, huang2023dynamic, chen2022towards, luo2024grasping, wu2024learning, li2024grasp, li2023gendexgrasp, wang2023dexgraspnet, liu2025parameterized, chen2024vividex, yin2023rotating, handa2023dextreme, lin2024twisting, li2024tpgp, oh2024bimanual, jiang2024dexmimicgen}. Achieving human-like proficiency in complex bimanual tasks holds significant research value and is crucial for progress toward general AI.

Thus, the rapid acquisition of precise, large-scale, and human-like dexterous manipulation sequences for data-driven embodied agents training~\cite{brohan2022rt, brohan2023rt, chi2023diffusion, o2023open, Ze2024DP3} becomes increasingly urgent. Some studies use reinforcement learning (RL)~\cite{kaelbling1996reinforcement, schulman2017proximal} to explore and generate dexterous hand actions~\cite{xu2023unidexgrasp, wan2023unidexgrasp++, zhang2024graspxl, luo2024grasping, christen2022d, zhang2024artigrasp, liu2023dexrepnet}, while others collect human-robot paired data through teleoperation~\cite{he2024learning, chi2024universal, wang2024dexcap, yang2024ace, he2024omnih2o, oh2024bimanual, shaw2024bimanual}. Both methods are limited: traditional RL requires carefully designed, task-specific reward functions~\cite{zhang2024graspxl, luo2024smplolympics}, restricting scalability and task complexity, while teleoperation is labor-intensive and costly, yielding only embodiment-specific datasets.

A promising solution is to transfer human manipulation actions to dexterous robotic hands in simulated environments via imitation learning~\cite{mandlekar2018roboturk, wang2023mimicplay, zhaodexh2r, liu2024force, qin2022dexmv}. This approach offers several advantages. First, imitating human manipulation trajectories creates naturalistic hand-object interactions, enabling more fluid and human-like motions. Second, abundant motion-capture (MoCap) datasets~\cite{garcia2018first, Brahmbhatt_2020_ECCV, taheri2020grab, hampali2020honnotate, chao2021dexycb, yang2022oakink, liu2022hoi4d, kwon2021h2o, xie2023hmdo, fan2023arctic, liu2024taco, Li_Yang_Lin_Xu_Zhan_Zhao_Zhu_Kang_Wu_Lu_2024, zhan2024oakink2} and hand pose estimation techniques~\cite{cao2021reconstructing, tekin2019h+, hasson2020leveraging, lin2021end, yang2021cpf, yang2022artiboost, xu2022vitpose, xu2023vitpose++, yang2023poem, pavlakos2024reconstructing} makes extracting operational knowledge from human demonstrations easily accessible~\cite{qin2022dexmv, shaw2023videodex}. Third, simulations provide a cost-effective validation, offering a shortcut to real-world robot deployment~\cite{jiang2024transic, he2024omnih2o, handa2023dextreme}.

Yet, achieving precise and efficient transfer is non-trivial. As shown in \cref{fig:teaser}, morphological differences between human and robotic hands lead to direct pose retargeting suboptimal. Additionally, although MoCap data is relatively accurate, error accumulation can still lead to critical failures during high-precision tasks\del{hindering accurate rollouts in physical simulations,}. Moreover, bimanual manipulation introduces a high-dimensional action space, significantly increasing the difficulty of efficient policy learning. Consequently, most pioneering work generally stops at single-hand grasping and lifting tasks~\cite{xu2023unidexgrasp, wan2023unidexgrasp++, zhang2024graspxl, christen2022d}, leaving complex bimanual activities—such as unscrewing a bottle or capping a pen—largely unexplored.

In this paper, we propose a simple but efficient method, \method, which facilitates the transfer of hand manipulation skills—especially bimanual actions—to dexterous robotic hands in simulation, enabling accurate tracking of reference motions. \textit{Our key insight is to treat the transfer as a two-stage process: a pre-training trajectory imitation stage focusing on hand motion alone, followed by a specific action fine-tuning stage that meets interaction constraints.} Specifically, we design a robust generalist model that learns to accurately mimic human finger motions with resilience to noise. Based on this initial imitation, we then introduce a residual learning module~\cite{johannink2019residual, silver2018residual, huang2024efficient, jiang2024transic} that incrementally refines the robot's actions, focusing on two key aspects: 1) ensuring stable contact with object surfaces under physical constraints, enabling effective object manipulation, and 2) coordinating both hands to ensure precise, high-fidelity execution of complex bimanual operations.

The advantages of this design are threefold: 1) In the first stage, focusing on dynamic hand mimicry with large-scale pretraining \textbf{effectively mitigates morphological differences}. 2) Building on this advantage, the second stage concentrates on tracking bimanual object interactions, \textbf{enabling precise capture of subtle movements and facilitating natural, high-fidelity manipulation}. 3) It \textbf{significantly reduces action space complexity} by decoupling human hand motion imitation from physics-based object interaction constraints, thus improving training efficiency.

Building on this framework, \method corrects arbitrary, noisy hand MoCap data into physically plausible motion without predefined stages (\eg, ``approaching-grasping-manipulation”) or task-specific reward engineering. We, therefore, validate its effectiveness and efficiency across a range of complex single- and bimanual manipulations, including articulated object handling~\cite{Li_Yang_Lin_Xu_Zhan_Zhao_Zhu_Kang_Wu_Lu_2024, zhan2024oakink2, taheri2020grab, fu2024gigahands, fan2023arctic}. Using \method, we transfer several representative hand-object manipulation datasets~\cite{Li_Yang_Lin_Xu_Zhan_Zhao_Zhu_Kang_Wu_Lu_2024, zhan2024oakink2} to dexterous robotic hands in the Isaac Gym simulation~\cite{makoviychuk2021isaac}, constructing the \dataset dataset, which achieves marked improvements in motion fidelity and compliance. Currently, \dataset comprises 3.3K episodes and 1.34 million frames of robotic hand manipulation, covering previously unexplored tasks such as pen capping, bottle cap unscrewing, and chemical experimentation.

We experimentally demonstrate that \method outperforms baseline methods in both motion precision and transfer success rate. Notably, it surpasses prior state-of-the-art (SOTA) approaches in transfer efficiency, even on a personal computer. To evaluate its extensibility, we conducted cross-embodiment experiments applying \method to dexterous hands with varying degrees of freedom (DoFs) and morphologies, achieving consistent performance with minimal additional effort.
Furthermore, we replay \dataset's bimanual trajectories on real-world devices, demonstrating agile and natural dexterous manipulation that, to the best of our knowledge, has not been achieved by previous RL- or teleoperation-based methods. Finally, we benchmark \dataset using several imitation learning frameworks, underscoring its value to the research community.

In summary, our contributions are as follows:

\begin{itemize}
    \item We introduce \method, a simple yet effective two-stage transfer framework that enables precise transfer of human bimanual manipulation to dexterous robotic hands in simulation, ensuring accurate tracking of both hand and object reference motions.
    \item Using this framework, we construct \dataset, a large-scale, high-quality dataset featuring a wide array of novel bimanual manipulation tasks with high precision and compliance. \dataset is extensible and serves as a valuable resource for future policy training.
    \item Our experiments show that \method outperforms previous SOTA methods. We further demonstrate its generalizability across various dexterous hand configurations and its feasibility for real-world deployment.
\end{itemize}
\section{Related Works}
\label{sec:related_works}

\qheading{Dexterous Manipulation via Human Demonstration}
Learning manipulation skills from human demonstrations offers an intuitive and effective approach to transferring human abilities to robots~\cite{argall2009survey, englert2018learning, zakka2023robopianist, ye2023learning}. Imitation learning has shown considerable promise in achieving this transfer~\cite{mandlekar2018roboturk, wang2023mimicplay, zhaodexh2r, liu2024force, tessler2024maskedmimic, peng2021amp, peng2022ase, zhou2024learning, li2023dexdeform, arunachalam2023dexterous, chen2022dextransfer}.
Recent studies focus on learning RL policies guided by object trajectories~\cite{chen2024vividex, luo2024grasping, zhou2024learning, chenobject, liu2025parameterized}. QuasiSim~\cite{liu2025parameterized} advances this approach by directly transferring reference hand motions to robotic hands via parameterized quasi-physical simulators. However, these methods are limited to simpler tasks and are computationally intensive. More recently, tailored solutions using task-specific reward functions have been developed for challenging tasks like bimanual lip-twisting~\cite{lin2024twisting, liu2024masked}. In contrast, our method enables efficient learning of complex manipulation tasks without task-specific reward engineering.

\qheading{Dexterous Hand Datasets}
Object manipulation is fundamental for embodied agents. Numerous MANO-based~\cite{MANO:SIGGRAPHASIA:2017} hand-object interaction datasets exist~\cite{garcia2018first, Brahmbhatt_2019_CVPR, Brahmbhatt_2020_ECCV, taheri2020grab, hampali2020honnotate, hampali2022keypoint, sener2022assembly101, chao2021dexycb, qin2022dexmv, yang2022oakink, liu2022hoi4d, kwon2021h2o, xie2023hmdo, fan2023arctic, zhu2023contactart, liu2024taco, Li_Yang_Lin_Xu_Zhan_Zhao_Zhu_Kang_Wu_Lu_2024, kim2024parahome, zhan2024oakink2, zhong2024color, hasson2019learning, gao2022dart, li2023chord, corona2020ganhand, li2024semgrasp}. However, these datasets often prioritize pose alignment with 2D images while neglecting physical constraints, limiting their applicability for robotic training.
Teleoperation methods~\cite{he2024learning, chi2024universal, wang2024dexcap, yang2024ace, he2024omnih2o, zhao2023learning, wu2023gello, qin2022one} collect human-to-robot hand matching data online using AR/VR systems~\cite{cheng2024open, ding2024bunny, chen2024arcap, jiang2024dexmimicgen, park2024dexhub} or vision-based MoCap~\cite{wang2024dexcap, qin2023anyteleop, wang2024cyberdemo} for real-time data acquisition and correction with humans in the loop. However, teleoperation is labor-intensive and time-consuming, and the absence of tactile feedback often yields stiff, unnatural actions, hindering fine-grained manipulation.
In contrast, our method enables offline transfer of human demonstrations to robots. Our \dataset offers a large, easily expandable collection of human demonstration episodes.

\qheading{Residual Learning}
Due to the sample inefficiency and time-consuming nature of RL training, residual policy learning~\cite{johannink2019residual, silver2018residual, schoettler2020deep}, which incrementally refines action control, is widely adopted to enhance efficiency and stability. In dexterous hand manipulation, various studies explore residual strategies tailored to specific tasks~\cite{alakuijala2021residual, davchev2022residual, schoettler2020deep, garcia2020physics, zhang2023efficient, wu2024learning, chenobject, zhaodexh2r}. For instance, \cite{garcia2020physics} integrates user input during residual policy training, while \cite{jiang2024transic} learns corrective actions from human demonstrations. GraspGF~\cite{wu2024learning} employs a pre-trained score-based generative model as a base, and \cite{chenobject} decomposes the imitation task into wrist following and finger motion control, integrating a residual wrist control policy. Additionally, \cite{huang2024efficient} constructs a mixture-of-experts system~\cite{jacobs1991adaptive} using residual learning, and DexH2R~\cite{zhaodexh2r} applies residual learning directly to retargeted robotic hand actions.
Our method differs from these approaches by pre-training a finger motion imitation model that incorporates additional dynamic information, followed by fine-tuning a residual policy to adapt to task-specific physical constraints. This approach is more efficient and generalizable across various manipulation tasks.
\section{Method}
\label{sec:method}

\begin{figure*}[t!]
    \centering
    \colorbox{white}{\includegraphics[width=0.95\textwidth]{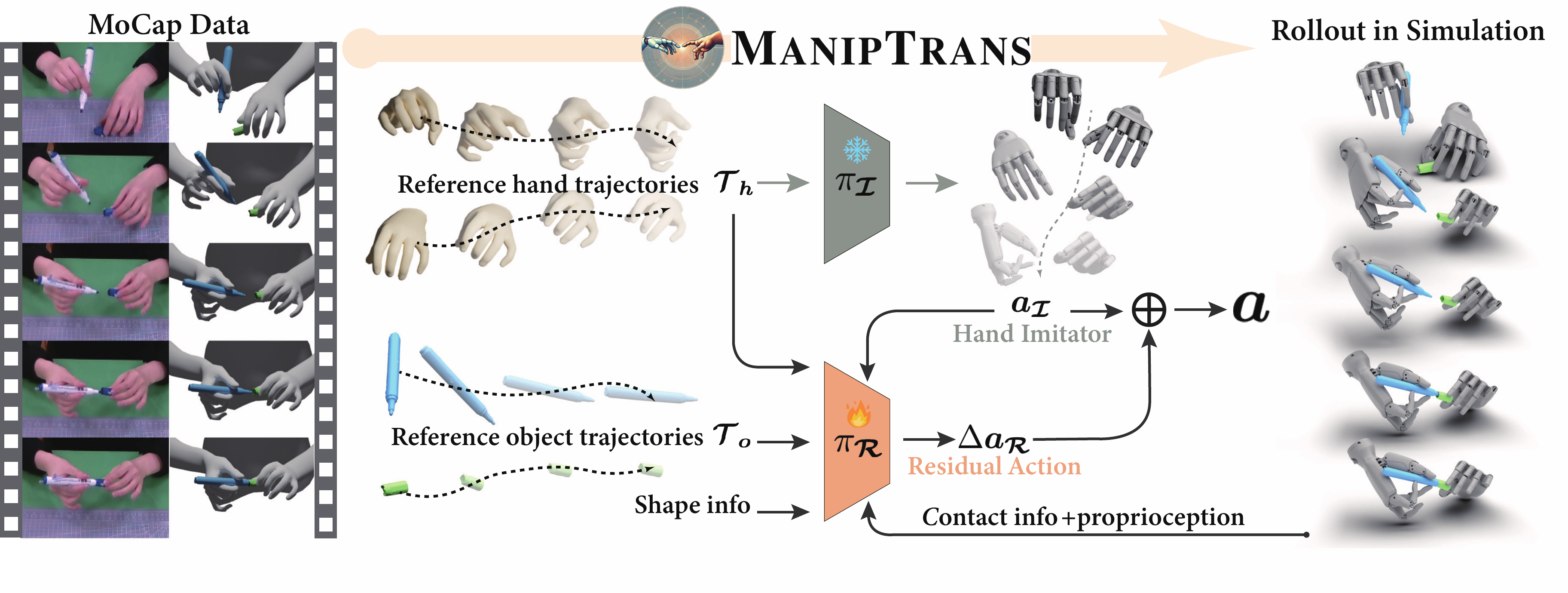}}
    \vspace{-2em}
    \caption{\textbf{Our \method Pipeline.} We first pre-train a hand motion imitation model with large-scale human demonstrations, then fine-tune a residual policy to adapt to task-specific physical constraints.
    }
    \vspace{-1em}
    \label{fig:pipeline}
\end{figure*}

We provide an overview of our method in \cref{fig:pipeline}. Given reference human hand–object interaction trajectories, our goal is to learn a policy that enables dexterous robotic hands to accurately replicate these trajectories in simulation while satisfying the task's semantic manipulation constraints. To this end, we propose a two-stage framework: the first stage trains a general hand trajectory imitation model, and the second stage employs a residual model to refine the initial coarse motion into task-compliant actions.

\subsection{Preliminaries}
\label{sec:method_preliminary}

Without loss of generality, we formulate the manipulation transfer problem in a complex bimanual setting, where the left and right dexterous hands, $\dexhand = \{d_l, d_r\}$, aim to replicate the behavior of human hands, $\hand = \{h_l, h_r\}$, which interact with two objects, $\obj = \{o_l, o_r\}$, in a cooperative manner (e.g., in a pen-capping task where one hand holds the cap while the other grips the pen body). The reference trajectories from human demonstrations are defined as $\Traj_{\hand} = \{\traj_{\hand}^t\}_{t=1}^{T}$ and $\Traj_{\obj} = \{\traj_{\obj}^t\}_{t=1}^{T}$, where $T$ represents the total number of frames.
The trajectory $\traj_{\hand}$ for each hand includes the wrist's 6-DoF pose $\mbold{w}_{\hand} \in \mathbb{SE}(3)$, the linear and angular velocities $\dot{\mbold{w}}_{\hand} = \{\mbold{v}_{\hand}, \mbold{u}_{\hand}\}$, and the finger joint positions $\mbold{j}_{\hand} \in \mathbb{R}^{F \times 3}$ defined by MANO~\cite{MANO:SIGGRAPHASIA:2017}, along with their respective velocities $\dot{\mbold{j}}_{\hand} = \{\mbold{v}_{\mbold{j}}, \mbold{u}_{\mbold{j}}\}$; here, $F$ denotes the number of hand keypoints, including the fingertips. Similarly, the object trajectory $\traj_{\obj}$ for each object includes its 6-DoF pose $\mbold{p}_{\obj} \in \mathbb{SE}(3)$ and the corresponding linear and angular velocities $\dot{\mbold{p}}_{\obj} = \{\mbold{v}_{\obj}, \mbold{u}_{\obj}\}$. To reduce spatial complexity, we normalize all translations relative to the dexterous hand's wrist position while preserving the original rotations to maintain the correct gravity direction.

We model this problem as an implicit Markov Decision Process (MDP) $\mathcal{M} = \langle \mbold{\mathcal{S}}, \mbold{\mathcal{A}}, \mbold{\mathsf{T}}, \Reward, \gamma\rangle$, where $\mbold{\mathcal{S}}$ represents the state space, $\mbold{\mathcal{A}}$ the action space, $\mbold{\mathsf{T}}$ the transition dynamics, $\Reward$ the reward function, and $\gamma$ the discount factor. The action for each dexterous hand at time $t$, denoted as $\mbold{a}^t \in \mbold{\mathcal{A}}$, comprises the target positions of each dexterous hand's joint $\mbold{a}_{\mbold{q}}^t \in \mathbb{R}^K$ for proportional-derivative (PD) control, and the 6-DoF force $\mbold{a}_{\mbold{w}}^t \in \mathbb{R}^6$ applied to the robotic wrist, similar to prior work~\cite{xu2023unidexgrasp, wan2023unidexgrasp++, huang2024efficient}, where $K$ denotes the total number of robotic hand revolute joints (\ie the DoF).

Our approach divides the transfer process into two stages: 1) a pre-trained hand-only trajectory imitation model $\imitator$, and 2) a residual module $\residual$ that fine-tunes the coarse actions to ensure task compliance. The state at time $t$ is defined separately for each stage as $\mbold{s}^t_{\imitator} \in \mbold{\mathcal{S}}_{\imitator}$ and $\mbold{s}^t_{\residual} \in \mbold{\mathcal{S}}_{\residual}$, with corresponding reward functions $r^t_{\imitator} = \Reward(\mbold{s}^t_{\imitator}, \mbold{a}^t_{\imitator})$ and $r^t_{\residual} = \Reward(\mbold{s}^t_{\residual}, \mbold{a}^t_{\residual})$ as described in \cref{sec:method_imitating} and \cref{sec:method_residual}. For both stages, we employ proximal policy optimization (PPO)~\cite{schulman2017proximal} to maximize the discounted reward $\mathbb{E}\left[\smash{\textstyle\sum_{t=1}^T} \gamma^{t - 1}r^{t}_\text{stage}\right]$, following previous methods~\cite{chen2022towards,peng2021amp}.

\subsection{Hand Trajectory Imitating}
\label{sec:method_imitating}

In this stage, our objective is to learn a general hand trajectory imitation model, $\imitator$, capable of accurately replicating detailed human finger motions. The state for each dexterous hand at time $t$ is defined as $\mbold{s}^t_{\imitator} = \{\traj_{\hand}^t, \mbold{s}^t_\text{prop}\}$, which includes the target hand trajectory $\traj_{\hand}^t$ and the current proprioception $\mbold{s}^t_\text{prop} = \{\mbold{q}_{\dexhand}^t, \dot{\mbold{q}}_{\dexhand}^t, \mbold{w}_{\dexhand}^t, \dot{\mbold{w}}_{\dexhand}^t\}$. Here, $\mbold{q}_{\dexhand}^t$ and $\mbold{w}_{\dexhand}^t$ denote the joint angles and wrist poses, respectively, along with their corresponding velocities. We aim to train the policy $\pi_{\imitator}(\mbold{a}^t | \mbold{s}_{\imitator}^t, \mbold{a}^{t - 1})$ using RL to determine the actions $\mbold{a}_{\imitator}^t$.

\qheading{Reward Functions.}
The reward function $r^t_{\imitator}$ is designed to encourage the dexterous hands to track the reference hand trajectory $\traj_{\hand}^t$ while ensuring stability and smoothness. It comprises three components: 1) \textit{Wrist tracking reward} $r^t_{\text{wrist}}$: This reward minimizes the difference: $\mbold{w}_{\dexhand}^t \ominus \mbold{w}_{\hand}^t$ and $\dot{\mbold{w}}_{\dexhand}^t - \dot{\mbold{w}}_{\hand}^t$, $\ominus$ denotes the difference in $\mathbb{SE}(3)$ space.
2) \textit{Finger imitation reward} $r^t_{\text{finger}}$: This component encourages the dexterous hand to closely follow the reference finger joint positions. We manually select $F$ finger keypoints on the dexterous hand corresponding to the MANO model, denoted as $\mbold{j}_{\dexhand}$. The weights $w_{f}$ and decay rates $\lambda_{f}$ are empirically set to emphasize the fingertips, particularly those of the thumb, index, and middle fingers. The parameters are in the \supp. This design helps mitigate the impact of morphological differences between human and robotic hands:
\begin{equation}
    r^t_{\text{finger}} = \textstyle\sum_{f=1}^F { w_{f} \cdot \exp{(-\lambda_{f} \|\mbold{j}_{{\dexhand_f}}^t - \mbold{j}_{{\hand_f}}^t\|\smash{_{\scriptscriptstyle 2}^{\scriptscriptstyle 2}})}}
\label{eq:finger_reward}
\end{equation}
3) \textit{Smoothness Reward} $r^t_{\text{smooth}}$: To alleviate jerky motions, we introduce a smoothness reward that penalizes the power exerted on each joint, defined as the element-wise product of joint velocities and torques, similar to the approach in~\cite{luo2023perpetual}.
The total reward is defined as: $r^t_{\imitator} = w_{\text{wrist}} \cdot r^t_{\text{wrist}} + w_{\text{finger}} \cdot r^t_{\text{finger}} + w_{\text{smooth}} \cdot r^t_{\text{smooth}}$.

\qheading{Training Strategy.}
Decoupling hand imitation from object interaction offers additional benefits; specifically, $\pi_{\imitator}$ does not require challenging-to-acquire manipulation data. We train the policy using hand-only datasets, including existing hand motion collections~\cite{zhan2024oakink2, Li_Yang_Lin_Xu_Zhan_Zhao_Zhu_Kang_Wu_Lu_2024, taheri2020grab, gao2022dart, zhang2017hand, zimmermann2019freihand, chao2021dexycb} and synthetic data generated via interpolation~\cite{shoemake1985animating}. To balance training data between the left and right hands, we mirror these datasets; training time and additional details are provided in the \supp.
For efficiency, we employ reference state initialization (RSI) and early termination~\cite{peng2021amp, peng2018deepmimic}. If the dexterous hand keypoints $\mbold{j}_{\dexhand}$ deviate beyond a threshold $\epsilon_{\text{finger}}$, the episode terminates early and resets to a randomly sampled MoCap state. We also utilize curriculum learning~\cite{bengio2009curriculum}, gradually reducing $\epsilon_{\text{finger}}$ to encourage broad exploration initially, then focusing on fine-grained finger control.


\subsection{Residual Learning for Interaction}
\label{sec:method_residual}

Building on the pre-trained $\pi_{\imitator}$, we use a residual module $\residual$ to refine coarse actions and satisfy task-specific constraints.

\qheading{State Space Expansion for Interaction.}
To account for interactions between the dexterous hands and objects, we expand the state space beyond the hand-related states $\mbold{s}^t_{\imitator}$ by incorporating additional interaction-related information.
First, we compute the convex hull~\cite{wei2022coacd} of the object meshes $\obj$ from MoCap data to generate the collidable object $\cobj$ in the simulation environment. To manipulate the object along the reference $\Traj_{\obj}$, we include the object's position $\mbold{p}_{\cobj}$ (relative to the wrist position $\mbold{w}_{\dexhand}$) and velocities $\dot{\mbold{p}}_{\cobj}$, center of mass $\mbold{m}_{\cobj}$, and gravitational force vector $\mbold{G}_{\cobj}$.
To better encode the object's shape, we utilize the BPS representation~\cite{prokudin2019efficient}. Additionally, for enhancing perception, we encode the spatial relationship between the hands and the object using the distance metric: $\mbold{D}(\mbold{j}_{\dexhand}^t, \mbold{p}_{\cobj}^t) = \textstyle{\|\mbold{j}_{{\dexhand}}^t - \mbold{p}_{{\cobj}}^t\|\smash{_{\scriptscriptstyle 2}^{\scriptscriptstyle 2}}}$, measuring the squared Euclidean distance between the dexterous hand keypoints and the object's position.
Furthermore, we explicitly include the contact force $\mbold{C}$ obtained from the simulation, capturing the interaction between the fingertips and the object's surface. This tactile feedback is critical for stable grasping and manipulation, ensuring precise control during complex tasks.
In summary, the expanded interaction state for the residual module is defined as: $\mbold{s}^t_{\text{interact}} = \{\traj_{\obj}^t, \mbold{p}_{\cobj}^t, \dot{\mbold{p}}_{\cobj}^t, \mbold{m}_{\cobj}^t, \mbold{G}_{\cobj}^t, \text{BPS}(\cobj), \mbold{D}(\mbold{j}_{\dexhand}^t, \mbold{p}_{\cobj}^t), \mbold{C}^t\}$.

\qheading{Residual Actions Combining Strategy.}
Given the combined state $\mbold{s}_{\residual}^t = \mbold{s}^t_{\imitator} \cup \mbold{s}^t_{\text{interact}}$, our goal is to learn residual actions $\Delta \mbold{a}_{\residual}^t$ that refine the initial imitation actions $\mbold{a}^t_{\imitator}$ to ensure task compliance. During each step of the manipulation episode, we first sample the imitation action $\mbold{a}^t_{\imitator} \sim \pi_{\imitator}(\mbold{a}^t | \mbold{s}_{\imitator}^t, \mbold{a}^{t - 1})$. Conditioned on this action, we then sample the residual correction $\Delta \mbold{a}_{\residual}^t \sim \pi_{\residual}(\Delta \mbold{a}^t | \mbold{s}_{\residual}^t, \mbold{a}^t_{\imitator}, \mbold{a}^{t - 1})$. The final action is computed as: $\mbold{a}^t = \mbold{a}^t_{\imitator} + \Delta \mbold{a}_{\residual}^t$, where the residual action is added element-wise. The resulting action $\mbold{a}^t$ is then clipped to adhere to the dexterous hand's joint limits.
At the start of training, since the dexterous hand movements already approximate the reference hand trajectory, the residual actions are expected to be close to zero. This initialization helps prevent model collapse and accelerates convergence. We achieve this by initializing the residual module with a zero-mean Gaussian distribution and employing a warm-up strategy to gradually activate its training.

\qheading{Reward Functions.}
Our objective is to efficiently transfer human bimanual manipulation skills to dexterous robotic hands in a task-agnostic manner. To this end, we avoid task-specific reward engineering, which, although beneficial for individual tasks, can limit generalization. Therefore, our reward design remains simple and general. In addition to the hand imitation reward $r^t_{\imitator}$ discussed in \cref{sec:method_imitating}, we introduce two additional components: 1) \textit{Object following reward} $r^t_{\text{object}}$: Minimizes positional and velocity differences between the simulated object and its reference trajectory, specifically $\mbold{p}_{\cobj}^t \ominus \mbold{p}_{\obj}^t$ and $\dot{\mbold{p}}_{\cobj}^t - \dot{\mbold{p}}_{\obj}^t$. 2) \textit{Contact force reward} $r^t_{\text{contact}}$: Encourages appropriate contact force when the hand-object distance in the MoCap dataset is below a specified threshold $\xi_{\text{c}}$. The reward is defined as:
\begin{equation}
    r^t_{\text{contact}} = w_{\text{c}} \cdot \exp{(\frac{-\lambda_{\text{c}}}{\textstyle\sum_{f=1}^F\mbold{C}_{\dexhand_f}^t \cdot \mathds{1}\left(\mbold{D}(\mbold{j}_{\hand_f}^t, \mbold{p}_{\obj}^t \cdot \obj) < \xi_{\text{c}}\right)})}
    \label{eq:contact_reward}
\end{equation}
where $\mbold{D}(\mbold{j}_{\hand_f}^t, \mbold{p}_{\obj}^t \cdot \obj)$ represents the minimum distance between the fingertip $\hand_f$ and the transformed object surface, $\mathds{1}(\cdot)$ is the indicator function, and $\mbold{C}_{\dexhand_f}^t$ denotes the contact force at the fingertip. The weight $w_{\text{c}}$ and decay rate $\lambda_{\text{c}}$ are empirically set to balance the reward function.
The total reward for the residual stage is defined as $r^t_{\residual} = r^t_{\imitator} + w_{\text{object}} \cdot r^t_{\text{object}} + w_{\text{contact}} \cdot r^t_{\text{contact}}$.

\qheading{Training Strategy.}
Inspired by prior work~\cite{pang2023global, pang2021convex, liu2025parameterized} that utilizes quasi-physical simulators to relax constraints during training and avoid local minima, we introduce a relaxation mechanism in the residual learning stage. Unlike \cite{liu2025parameterized}, which employs custom simulations, we adjust the physical constraints directly within the Isaac Gym environment~\cite{makoviychuk2021isaac} to enhance training efficiency.
Specifically, we initially set the gravitational constant $\mathcal{G}$ to zero and the friction coefficient $\mathcal{F}$ to a high value. This setup allows the robotic hands to, early in training, grip objects firmly and efficiently align with reference trajectories.
As training progresses, we gradually restore $\mathcal{G}$ to its true value and reduce $\mathcal{F}$ to a suitable value to approximate real interactions.
Similar to the imitation stage, we adopt RSI, early termination, and curriculum learning strategies. Each episode initializes the robotic hands by randomly selecting a non-colliding near-object state from the preprocessed trajectory. During training, if the object's pose $\mbold{p}_{\cobj}^t$ deviates beyond a predefined threshold $\epsilon_{\text{object}}$, the episode is terminated early. We progressively reduce $\epsilon_{\text{object}}$ to encourage more precise object manipulation.
Additionally, we introduce a contact termination condition: if MoCap data indicates a firm grasp by the human hands (i.e., $\mbold{D}(\mbold{j}_{\hand_f}^t, \mbold{p}_{\obj}^t \cdot \obj) < \xi_{\text{t}}$, where $\xi_{\text{t}}$ is the termination threshold), the contact force $\mbold{C}_{\dexhand_f}^t$ must be non-zero. Failure to meet this condition results in early termination. This mechanism ensures the agent learns to control contact forces, promoting stable object manipulation.

\subsection{\dataset Dataset}

Using \method, we generate \dataset, derived from two representative large-scale hand-object interaction datasets: FAVOR~\cite{Li_Yang_Lin_Xu_Zhan_Zhao_Zhu_Kang_Wu_Lu_2024} and OakInk-V2~\cite{zhan2024oakink2}. FAVOR employs VR-based teleoperation with human-in-the-loop corrections, focusing on foundational tasks like object rearrangement. In contrast, OakInk-V2 utilizes optical tracking-based motion capture, targeting more complex interactions such as pen capping and bottle unscrewing.

Due to the lack of standardization in dexterous robotic hands, we adopt the Inspire Hand~\cite{inspirehandurl} as our primary platform for its high dexterity, stability, cost-effectiveness, and extensive prior use~\cite{jiang2024dexmimicgen, fu2024humanplus, cheng2024open}.
To address the complexity of bimanual tasks, we employ a simulated 12-DoF configuration of the Inspire Hand, enhancing flexibility compared to its real-world 6-DoF mechanism.\del{ For real-world deployment, we adapt the robot based on optimization.} We demonstrate \method's adaptability to other robotic hands and real-world deployment in \cref{sec:cross_embodied} and \cref{sec:real_world_deployment}.

Our \dataset encompasses 61 diverse and challenging tasks as defined in~\cite{zhan2024oakink2}, comprising 3.3K episodes of robotic hand manipulation over 1.2K objects, totaling 1.34 million frames, including $\sim600$ sequences involving complex bimanual tasks. Each episode executes precisely in the Isaac Gym simulation~\cite{makoviychuk2021isaac}. In comparison, a recent dataset generated via automated augmentation~\cite{jiang2024dexmimicgen} includes only 60 source human demonstrations across 9 tasks.

\section{Experiments}
\label{sec:experiments}

In experiments, we describe the dataset setup and metrics (\cref{sec:dataset_metrics}), followed by implementation details (\cref{sec:implementation_details}). We then compare \method with SOTA methods (\cref{sec:comparisons}), demonstrate cross-embodiment generalization (\cref{sec:cross_embodied}), validate real-world deployment (\cref{sec:real_world_deployment}), conduct ablation studies (\cref{sec:ablation_study}), and benchmark \dataset for learning manipulation policies (\cref{sec:dataset_evaluation}).

\subsection{Datasets and Metrics}
\label{sec:dataset_metrics}

\qheading{Datasets}
For quantitative evaluation, we use the official validation dataset of OakInk-V2~\cite{zhan2024oakink2}, approximately half of which consists of bimanual tasks. To assess transfer capabilities, we manually select MoCap sequences that meet task completeness and semantic relevance, filtering them to durations of 4–20 seconds and downsampling to 60 fps. We exclude sequences involving deformable or oversized objects, resulting in $\sim80$ episodes. For qualitative evaluation, we also incorporate the GRAB~\cite{taheri2020grab}, FAOVR~\cite{Li_Yang_Lin_Xu_Zhan_Zhao_Zhu_Kang_Wu_Lu_2024}, and ARCTIC~\cite{fan2023arctic} datasets to demonstrate our advantages.

\qheading{Metrics} To evaluate \method in terms of manipulation precision, task compliance, and transfer efficiency, we introduce the following metrics. These are adapted from~\cite{liu2025parameterized} but are more stringent due to the complexity of our bimanual tasks: 1) Per-frame Average Object Rotation and Translation Error:$E_{r} = \frac{1}{T}\textstyle\sum_{t=1}^T({\mbold{p}_{\text{rot}}}_{\cobj}^t \cdot {({\mbold{p}_{\text{rot}}}_{\obj}^t)}^{-1})$ and $E_{t} = \frac{1}{T}\textstyle\sum_{t=1}^T\|{\mbold{p}_{\text{tsl}}}_{\cobj}^t - {{\mbold{p}_{\text{tsl}}}_{\obj}^t}\|\smash{_{\scriptscriptstyle 2}^{\scriptscriptstyle 2}}$. Here, $\mbold{p}_{\text{rot}}$ and $\mbold{p}_{\text{tsl}}$ are the rotation and translation components of the 6-DoF pose $\mbold{p}$, respectively. Errors $E_{r}$ and $E_{t}$ are reported in degrees and centimeters.
2) Mean Per-Joint Position Error (in $cm$): $E_{j} = \frac{1}{T \cdot F}\textstyle\sum_{t=1}^T \textstyle\sum_{f=1}^F \|\mbold{j}_{{\dexhand_f}}^t - \mbold{j}_{{\hand_f}}^t\|\smash{_{\scriptscriptstyle 2}^{\scriptscriptstyle 2}}$. This metric measures the average error in the positions of the hand joints.
3) Mean Per-Fingertip Position Error (in $cm$): $E_{ft} = \frac{1}{T\cdot M}\textstyle\sum_{t=1}^T \textstyle\sum_{ft=1}^{M} \|\mbold{t}_{{\dexhand_{ft}}}^t - \mbold{t}_{{\hand_{ft}}}^t\|\smash{_{\scriptscriptstyle 2}^{\scriptscriptstyle 2}}$.
This metric evaluates the mimicry quality of fingertip $\mbold{t}$ motions, accounting for morphological differences between human and robotic hands. Here, $M$ equals 5 for single-hand tasks and 10 for bimanual tasks.
4) Success Rate ($SR$): A tracking attempt is deemed successful if $E_{r}$, $E_{t}$, $E_{j}$, and $E_{ft}$ are all below the specified thresholds: $30^\circ$, $3\ cm$, $8\ cm$, and $6\ cm$, respectively. For bimanual tasks, the trajectory is considered failed if either hand fails to meet these conditions, making the success criterion stricter compared to single-hand tasks.

\subsection{Implementation Details}
\label{sec:implementation_details}
In \method, we manually selected $F = 21$ keypoints on each dexterous robotic hand, corresponding to the fingertips, palm, and phalangeal positions on the human hand, to mitigate the morphological differences. Details on keypoint selection and weight coefficients $w$ for reward terms are provided in \supp.
For training, we use a curriculum learning strategy. The initial threshold $\epsilon_{\text{finger}}$ is set to $6\ cm$ and decays to $4\ cm$. Object alignment thresholds $\epsilon_\text{object}$ start at $90^\circ$ and $6\ cm$ for rotation and translation, gradually decreasing to $30^\circ$ and $2 \ cm$.
We train both the imitation module $\imitator$ and residual module $\residual$ using the Actor-Critic PPO algorithm~\cite{schulman2017proximal}, with a training horizon of 32 frames, a mini-batch size of 1024, and a discount factor $\gamma = 0.99$. Optimization employs Adam~\cite{KingBa15} with an initial learning rate of $5 \times 10^{-4}$ and a decay scheduler.
All experiments are run in Isaac Gym~\cite{makoviychuk2021isaac}, simulating 4096 environments at a time step of $1/60$ s on a personal computer equipped with an NVIDIA RTX 4090 GPU and an Intel i9-13900KF CPU.

\subsection{Evaluations}
\label{sec:comparisons}

As discussed in \cref{sec:related_works}, dexterous hand manipulation advances rapidly, with previous approaches differing in problem formulations and task definitions. To offer a comprehensive and fair comparison, we evaluate two categories of methods—RL-combined and optimization-based—to demonstrate \method's accuracy and efficiency.

\begin{figure*}[t!]
    \begin{center}
       \colorbox{white}{\includegraphics[width=0.95 \linewidth]{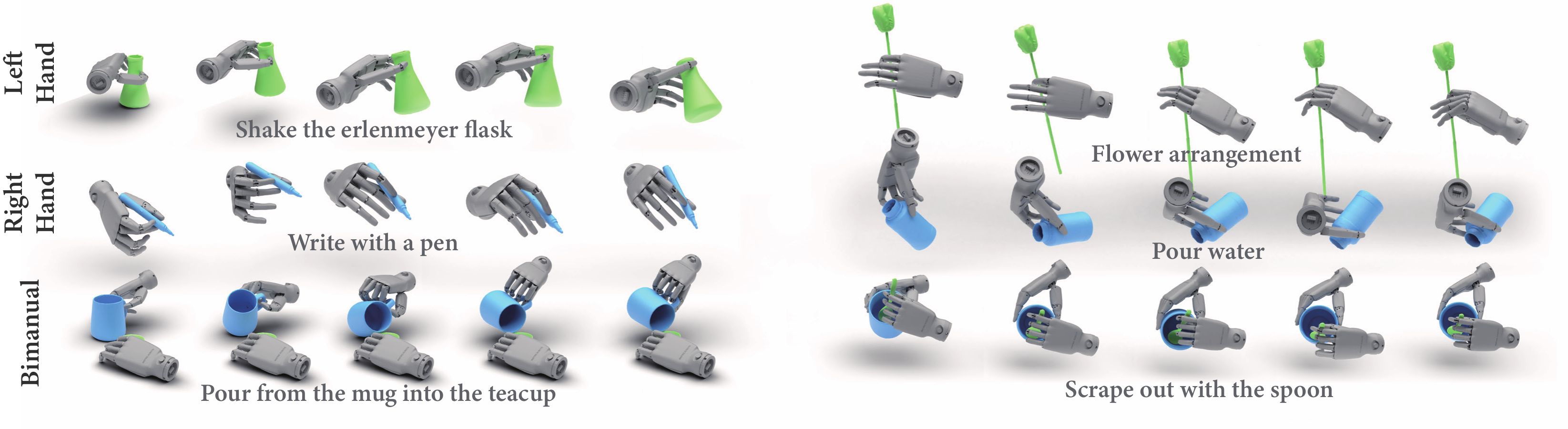}}
       \vspace{-0.8em}
       \caption{\textbf{Qualitative Results of \method.} We showcase the transfer results using the Inspire left and right hands on both single-hand tasks (top two rows) and bimanual tasks (bottom row) from the OakInk-V2~\cite{zhan2024oakink2} dataset. Notably, the dexterous hands successfully manipulate delicate and slim objects, such as a pen and a flower stem.}    
    \label{fig:qualitative_real_world_img}
    \vspace{-2.0em}
    \end{center}
\end{figure*}

\begin{table}[b]
    \rowcolors{2}{WhiteGray}{White}
    \renewcommand{\arraystretch}{1}
    \begin{center}
    \vspace{-1.0em}
    \resizebox{1.0\linewidth}{!}
    {
        \setlength{\tabcolsep}{9pt}
        {
        \begin{tabular}{cccccc}
        \toprule
        Methods & $E_{r} \downarrow$ & $E_{t} \downarrow$ & $E_{j} \downarrow$ & $E_{ft} \downarrow$ &$SR \uparrow$ \\
        \midrule
         \textit{Retarget-Only} & N/A & N/A & N/A & N/A & 4.6 / 0.0 \\
         \textit{RL-Only} & 9.72 & 1.23 & 2.96 &  2.38 & 34.3 / 12.1\\
         \textit{Retarget + Residual} & 11.58 & 0.79 & 2.54 & 1.74 & 47.8 / 13.9 \\
         \method & \textbf{8.60} & \textbf{0.49}& \textbf{2.15} & \textbf{1.36} & \textbf{58.1 / 39.5} \\
         \bottomrule
        \end{tabular}
        }
    }
    \caption{\textbf{Quantitative Comparisons with RL-Combined Baselines.} The first four metrics are computed only on successfully rolled-out sequences. The $SR$ includes the separated transfer success rates for single/bimanual tasks. The error scores on \textit{Retarget-Only} are not available since it hardly works.}
    \vspace{-1em}
    \label{tab:meth_cmp}
    \end{center}
\end{table}

\qheading{Comparison with RL-Combined Methods}
Due to the lack of publicly available code for prior RL-combined methods, we reimplement representative approaches: 1) \textit{RL-Only} exploration using only trajectory-following rewards, employing the PPO algorithm to train the robotic hand from scratch based on \cite{christen2022d}; 2) \textit{Retarget + Residual} learning, applying residual action to retargeted robotic hand poses obtained via alignment between human and robot keypoints~\cite{qin2023anyteleop}. As a naive baseline, we also include the \textit{Retarget-Only} method—retargeting without any learning.

As shown in \cref{tab:meth_cmp}, our method outperforms all baselines across multiple metrics, demonstrating superior precision in both single- and bimanual tasks. These results confirm that our two-stage transfer framework effectively captures subtle finger motions and object interactions, leading to high task success rates and motion fidelity.

We find that the \textit{Retarget-Only} baseline is nearly infeasible due to the complexity of the dexterous hand action space and error accumulation. The \textit{RL-Only} baseline performs suboptimally since exploration from scratch is time-consuming and reduces motion precision. Compared to the \textit{Retarget + Residual} baseline, our method—leveraging a pre-trained hand imitation model—demonstrates improved control capabilities, enabling more accurate manipulation aligned with the reference trajectory.
Notably, the Retargeting method often causes collisions in contact-rich scenarios, resulting in instability during residual policy training.
We further study \method's robustness and time cost in \supp. \cref{fig:qualitative_real_world_img} shows the qualitative results on seldom-explored tasks, highlighting the natural and precision of \method transferring human manipulation skills. Additional details and more qualitative results applying our method to articulated objects are provided in \supp.

\qheading{Comparison with Optimization-Based Method}
QuasiSim~\cite{liu2025parameterized} optimizes over customized simulations to track human motions. Currently, their full pipeline has not yet been released, and their ``randomly" selected validation set is not available. Thus, a direct quantitative comparison is not feasible. Therefore, we provide a qualitative comparison in \cref{fig:cmp_quasisim}, demonstrating \method's ability to transfer human motions to the Shadow Hand in a setting similar to QuasiSim's, but with more stable contacts and smoother motions.
Notably, due to our two-stage design, for an unseen single-hand manipulation trajectory of 60 frames (``rotating a mouse"), our method requires $\sim15$ minutes of training to achieve robust results, compared to QuasiSim's $\sim40$ hours of optimization\footnote{Results shown in QuasiSim's \supp and its official repository: \url{https://github.com/Meowuu7/QuasiSim}}, highlighting \method's significant efficiency.

\subsection{Cross-Embodiments Validation}
\label{sec:cross_embodied}

\begin{figure}[t!]
    \centering
    \colorbox{white}{\includegraphics[width=\linewidth]{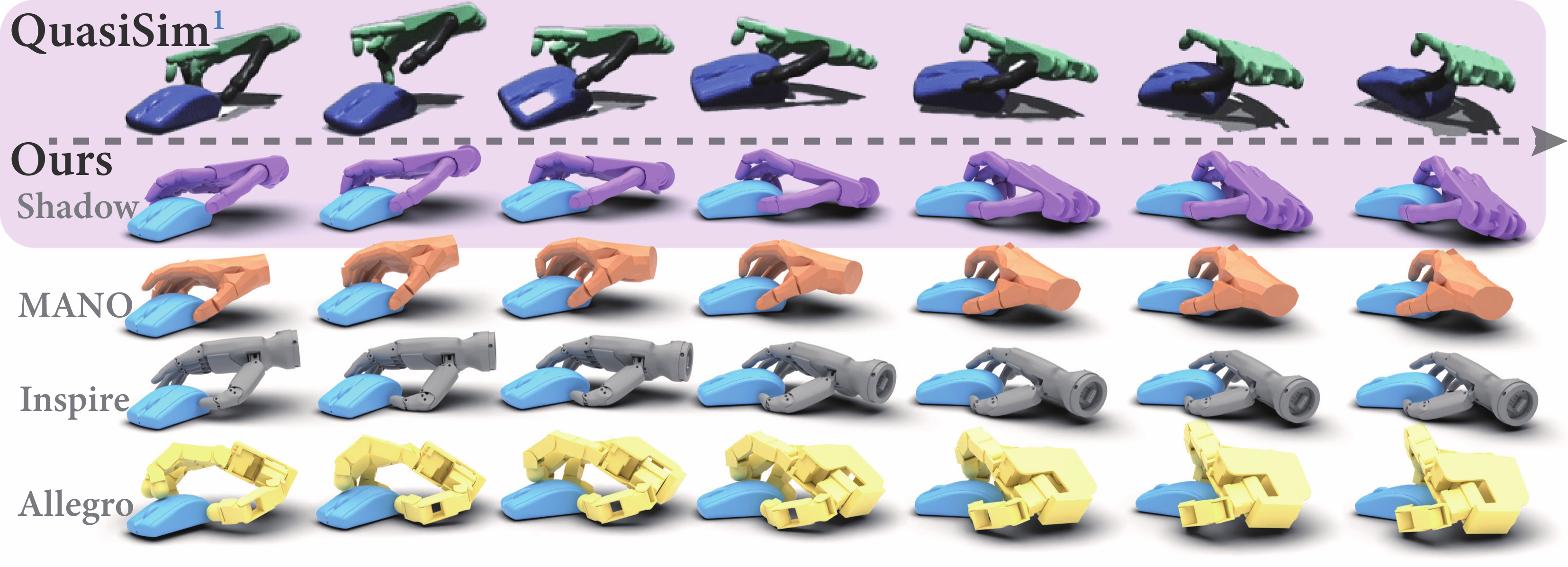}}
    \vspace{-1.5em}
    \caption{\textbf{Qualitative Comparison with QuasiSim~\cite{liu2025parameterized}.} \method produces more natural motion of the Shadow hand ({\color[RGB]{186,141,226}purple region}) and is applicable to other dexterous hands.}
    \label{fig:cmp_quasisim} 
    \vspace{-1.0em}
\end{figure}

\begin{figure}[t!]
    \centering
    \colorbox{white}{\includegraphics[width=\linewidth]{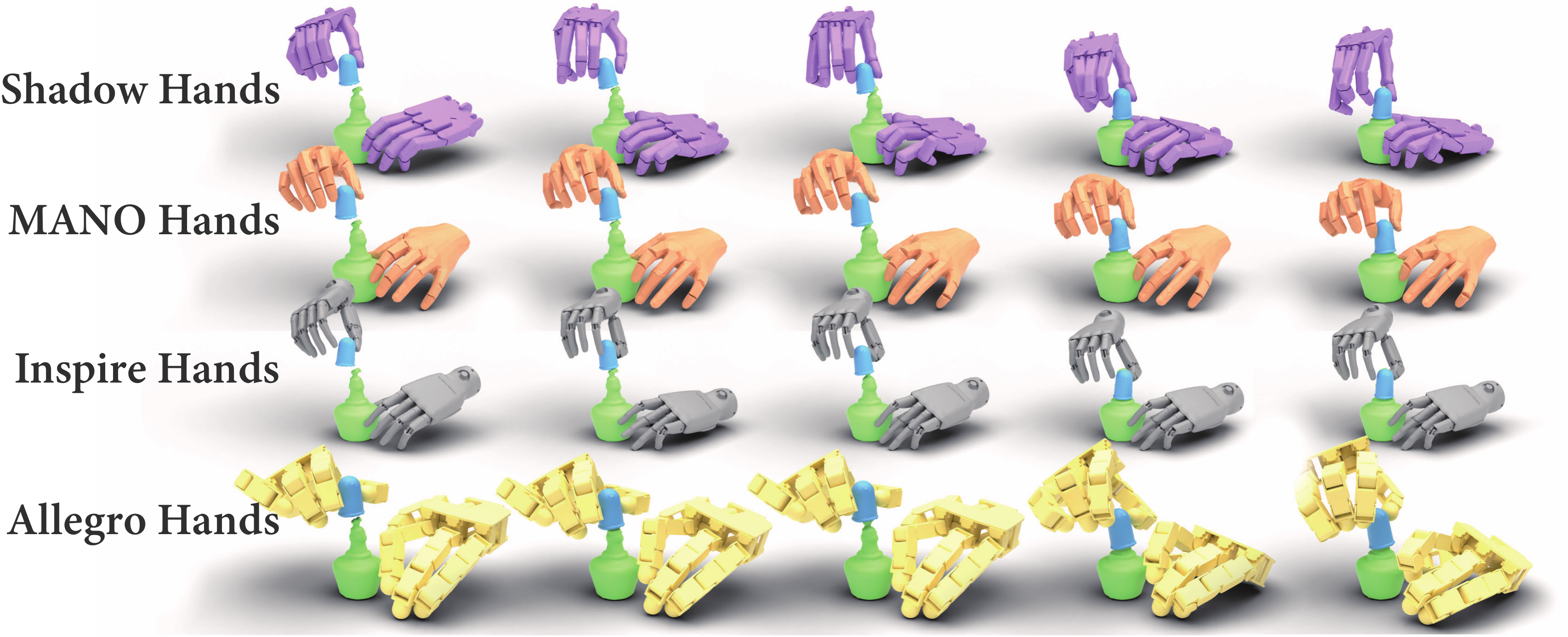}}
    \caption{\textbf{Cross Embodiments Results}: Putting off Alcohol lamp.}
    \vspace{-1em}
    \label{fig:cross_embodied}
\end{figure}

We demonstrate \method's extensibility across various dexterous hand embodiments. As described in \cref{sec:method}, the imitation module $\imitator$ addresses hand keypoint tracking, while the residual module $\residual$ captures physical interactions between fingertips and objects. Our framework is embodiment-agnostic since it relies solely on the correspondence between human fingers and robotic joints, allowing adaptation to different dexterous hands with minimal effort.
We evaluate \method on the Shadow Hand~\cite{shadowhandurl}, articulated MANO hand~\cite{MANO:SIGGRAPHASIA:2017, christen2022d}, Inspire Hand~\cite{inspirehandurl}, and Allegro Hand~\cite{allegrohandurl}, which have varying DoFs: $K = 22$, 22, 12, and 16, respectively. Without altering network hyperparameters or reward weights, \method achieves consistent, fluid, and precise performance across all embodiments in both single-hand tasks (\cref{fig:cmp_quasisim}) and bimanual tasks (\cref{fig:cross_embodied}). Additional details on the Allegro Hand—a robotic hand with only four fingers—are provided in \supp.

\subsection{Real-World Deployment}
\label{sec:real_world_deployment}

\begin{figure}[t]
    \centering
    \colorbox{white}{\includegraphics[width=\linewidth]{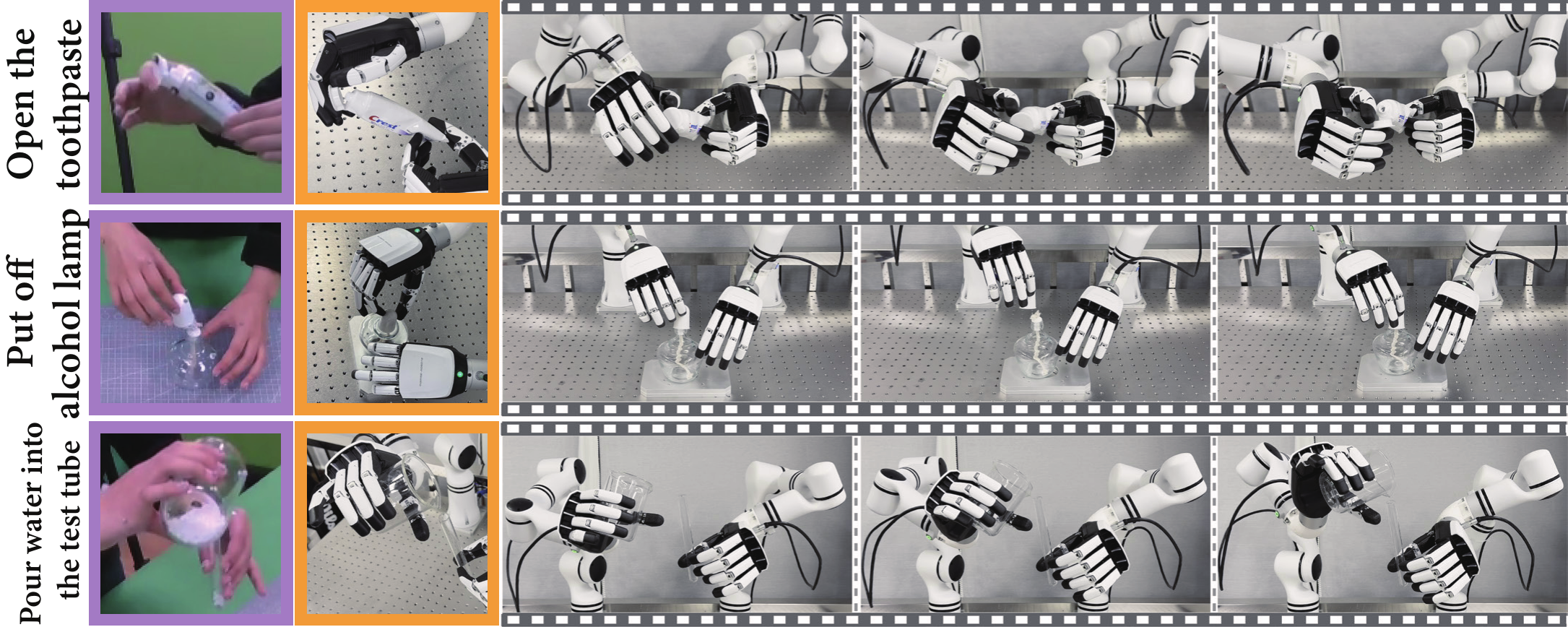}}
    \caption{\textbf{Real-world bimanual manipulation deployment.} {\color[RGB]{186,141,226}Purple} box: human hand motion; {\color[RGB]{251,176,59}orange} box: close-up of dexterous hands. More results are on the website. (Zoom in for details. {\color[RGB]{102,102,102}{\faSearchPlus}})}
    \label{fig:real_hardware}
    \vspace{-0.8em}
\end{figure}

As illustrated in \cref{fig:real_hardware}, we conduct experiments using two 7-DoF Realman arms~\cite{realmanarm} and a pair of upgraded Inspire Hands (same configuration yet adding tactile sensors). To bridge the gap between the simulated 12-DoF robotic hands and the 6-DoF real hardware, we employ a fitting-based method that optimizes the joint angles $\mbold{q}_{\tilde{\dexhand}} \in \mathbb{R}^{6}$ of the real robots (denoted as $\tilde{\cdot}$) for fingertip alignment, formulated as: $\operatorname*{argmin}_{\mbold{q}_{\tilde{\dexhand}}} \frac{1}{T\cdot M}\textstyle\sum_{t=1}^T \textstyle\sum_{ft=1}^{M} \|\mbold{t}_{{\dexhand_{ft}}}^t - \mbold{t}_{{\tilde{\dexhand}_{ft}}}^t\|\smash{_{\scriptscriptstyle 2}^{\scriptscriptstyle 2}}$ with an additional temporal smoothness loss: $L_\text{smooth} = \frac{1}{T-1}\textstyle\sum_{t=1}^{T-1} \|\mbold{q}_{\tilde{\dexhand}}^{t+1} - \mbold{q}_{\tilde{\dexhand}}^t\|\smash{_{\scriptscriptstyle 2}^{\scriptscriptstyle 2}}$.
We control the arms by solving inverse kinematics to align the arms' flanges with the dexterous hands' wrists $\mbold{w}_{\dexhand}$. During replay, we do not enforce strict temporal alignment, as the real robots cannot always operate as quickly as human hands.

\cref{fig:real_hardware} showcases dexterous manipulation that, to the best of our knowledge, has not previously been achieved. For example, in ``opening the toothpaste", the left hand stably holds the tube while the right hand's thumb and index finger flexibly pop open the tiny cap—motions challenging to capture via teleoperation. This underscores the potential of our method for future real-world policy learning.

\subsection{Abalation Studies}
\label{sec:ablation_study}

\begin{figure}[tb]
    \centering
    \begin{subfigure}{0.48\linewidth}
      \colorbox{white}{\includegraphics[width=\linewidth]{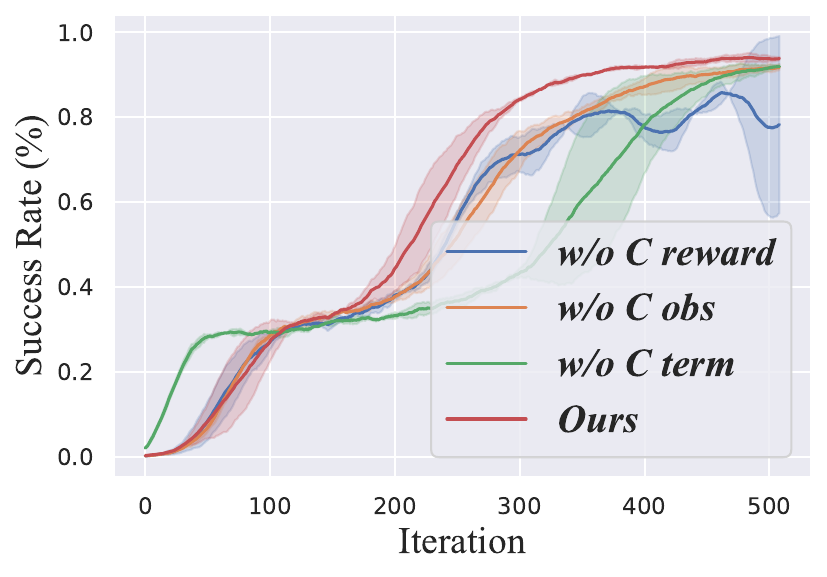}}
      \caption{Tactile ablations training curve.}
      \vspace{-0.2em}
      \label{fig:basemodel_cmp}
    \end{subfigure}
    \hfill
    \begin{subfigure}{0.48\linewidth}
      \colorbox{white}{\includegraphics[width=\linewidth]{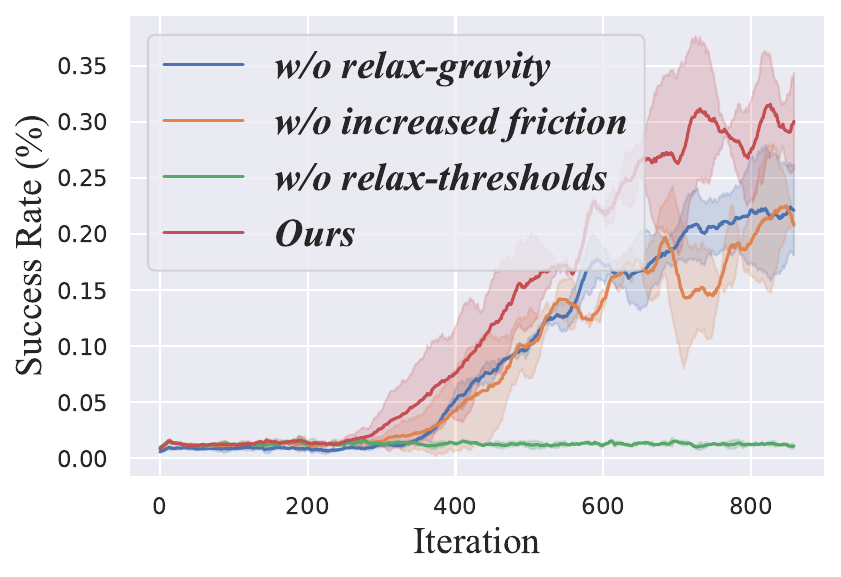}}
      \caption{Curve on training strategies.}
      \vspace{-0.2em}
      \label{fig:release_cmp}
    \end{subfigure}
    \caption{\textbf{Training Curve of Ablation Studies.} We assess tactile feedback in contact-rich tasks (\eg, turning off a lamp) and curriculum learning in complex ones (\eg, capping a pen).}
    \vspace{-0.28em}
\end{figure}

\begin{table}[t!]
    \rowcolors{4}{WhiteGray}{White}
    \renewcommand{\arraystretch}{1}
    \begin{center}
    \resizebox{1.0\linewidth}{!}
    {
        \setlength{\tabcolsep}{9pt}
        {
        \begin{tabular}{c|cccc}
            \toprule
            \textbf{Methods} & IBC~\cite{florence2022implicit} & BET~\cite{shafiullah2022behavior} & DP-UNet~\cite{chi2023diffusion} & DP-Trans~\cite{chi2023diffusion} \\
            \midrule
            $SR$   & 4.69\% & 9.69\% & \textbf{18.44\%} & 14.69\% \\
            \bottomrule
        \end{tabular}
        }
    }
    \vspace{-0.3em}
    \caption{\textbf{Imitating Learning on Bottle Rearrangement Task.}}
    \vspace{-2.5em}
    \label{tab:dataset_benchmark}
    \end{center}
\end{table}

\qheading{Tactile Information as Auxiliary Input}
In \cref{sec:method_residual}, we integrate tactile information, specifically the contact force $\mbold{C}$, into the pipeline in three ways: (1) as an observation input, (2) as a reward component to encourage contact, and (3) as a condition for early termination. Ablation studies (\cref{fig:basemodel_cmp}) labeled \textit{w/o $\mbold{C}$ obs}, \textit{w/o $\mbold{C}$ reward}, and \textit{w/o $\mbold{C}$ term} demonstrate that including $\mbold{C}$ in the reward function improves task success rates, and treating $\mbold{C}$ as an observation accelerates convergence. We also find that omitting $\mbold{C}$ as a termination condition seems to enhance initial training performance but lowers overall convergence speed, highlighting the importance of stable contact in task completion.

\qheading{Training Strategy}
We begin training with a curriculum learning strategy that includes (1) relaxing gravity effects, (2) increasing friction influence, and (3) relaxing thresholds $\epsilon_{\text{finger}}$ and $\epsilon_{\text{object}}$. Ablation studies (\cref{fig:release_cmp}), labeled \textit{w/o relax-gravity}, \textit{w/o increased friction}, and \textit{w/o relax-thresholds}, show that for precise, complex bimanual motions, ignoring gravity and using high friction coefficients in the early stages accelerate convergence and achieve higher overall $SR$. Without initial relaxation of the threshold constraints, the network may fail to converge entirely.

\subsection{\dataset for Policy Learning}
\label{sec:dataset_evaluation}

To benchmark \dataset's potential, we evaluate representative imitation learning methods on a fundamental policy learning task: rearrangement. Specifically, we focus on \textit{moving a bottle to a goal position}. Given the bottle's current and goal 6D poses, the environment state (including obstacles on the table), and the dexterous hand's proprioception, the policy generates a sequence of robotic hand actions to pick up the bottle and place it at the target.

We evaluate four representative imitation learning methods: two regression-based behavior cloning approaches—IBC~\cite{florence2022implicit} and BET~\cite{shafiullah2022behavior}—and two diffusion policy methods~\cite{chi2023diffusion} with UNet~\cite{ronneberger2015u} and Transformer~\cite{vaswani2017attention} backbones. Each policy is trained on 85\% of the 140 sequences involving the bottle rearrangement task in \dataset and evaluated on the remaining 15\%. We perform 20 rollouts per sequence. A rollout is considered successful if the object's final position is within $10~cm$ of the goal. Further details are provided in \supp.

As shown in \cref{tab:dataset_benchmark}, all methods perform suboptimally due to the task's difficulty and the complexity of the dexterous hand action space. Regression-based behavior cloning approaches, in particular, suffer from error accumulation. These results highlight the inherent challenges of dexterous manipulation tasks, which require precise finger control and effective object manipulation. We hope that \dataset will facilitate advancements in this domain.

\section{Conclusion and Discussion}
\label{sec:conclusion}

\method is a two-stage framework that efficiently transfers human manipulation skills to dexterous robotic hands. By decoupling hand motion imitation from object interaction via residual learning, \method overcomes morphological differences and complex task challenges, ensuring high-fidelity motions and efficient training. Experiments demonstrate that \method surpasses SOTA methods in motion precision and computational efficiency, while also exhibiting cross-embodiment adaptability and feasibility for real-world deployment. Furthermore, the extensible \dataset establishes a new benchmark to advance progress in embodied AI.

\qheading{Discussion and Limitations}
Although \method successfully handles most MoCap data, some sequences cannot be transferred effectively. We attribute this to two main reasons: 1) excessive noise in interaction poses and 2) insufficiently accurate object models for simulation, particularly for articulated objects. Enhancing \method's robustness and generating physically plausible object models are valuable directions for future research.

{
    \small
    \bibliographystyle{ieeenat_fullname}
    \bibliography{main}
}
\begin{appendix}
  \clearpage
\setcounter{page}{1}
\setcounter{footnote}{0}
\maketitlesupplementary

This appendix provides additional details and results that complement the main paper. We first validate the extensibility of \method in \cref{sec:supp_extension}. We then evaluate the robustness of \method under noisy conditions in \cref{sec:supp_robustness} and analyze its time cost in \cref{sec:supp_timecost}. Detailed information on the settings of \method is provided in \cref{sec:supp_details_method}, along with statistics for the \dataset dataset in \cref{sec:supp_dataset_stats}. Finally, we present the training details for the rearrangement policies in \cref{sec:supp_imitation}.

\section{Further Extension of \method}
\label{sec:supp_extension}

\subsection{Articulated Object Manipulation}

We demonstrate the extensibility of \method by applying it to the ARCTIC dataset~\cite{fan2023arctic}, which includes approximately 10 articulated objects, each with precise hand manipulation trajectories for bimanual single-object manipulation tasks.

To accommodate the articulated object manipulation task, we extend our method pipeline. For a single articulated object $\aobj$, we define its trajectory as $\Traj_{\aobj} = \{\traj^t_{\aobj}\}_{t=1}^{T}$, where $\traj_{\aobj} = \{\mbold{p}_{\aobj}, \dot{\mbold{p}}_{\aobj}, \theta_{\aobj}, \dot{\theta}_{\aobj}\}$ represents the object's transformation, velocity, and the angle and angular velocity of its articulated part. The reward function for articulated objects, $r^t_{\text{object}^\text{A}}$, includes two additional terms compared to the reward for rigid objects: the angle difference $|\theta_{\aobj} - \theta_{\caobj}|$ and the angular velocity difference $|\dot{\theta}_{\aobj} - \dot{\theta}_{\caobj}|$, where $\caobj$ represents the collidable articulated object in the simulation environment~\cite{makoviychuk2021isaac}. Apart from this modification, the rest of the pipeline remains unchanged.

\begin{figure}[t!]
    \centering
    \colorbox{white}{\includegraphics[width=\linewidth]{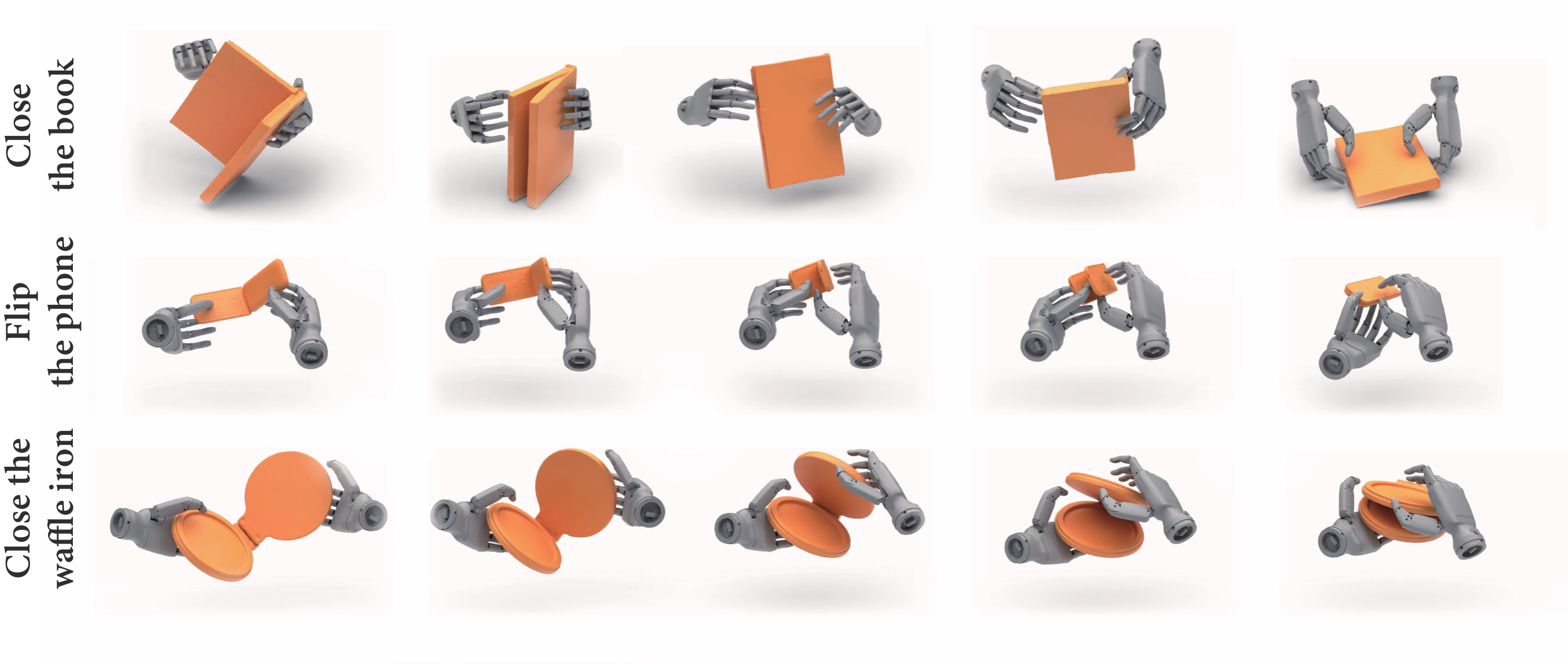}}
    \caption{\textbf{Applying \method to Articulated Object Manipulation.} In the first row, the two hands collaborate to not only close the book but also place it stably on the table.}
    \label{fig:supp_articulated}
\end{figure}

Qualitative results of \method applied to the ARCTIC dataset are presented in \cref{fig:supp_articulated}, demonstrating that our method successfully imitates human demonstrations and rotates the articulated object to the desired target angle. This highlights the extensibility of our pipeline when the physical properties of the articulated object can be accurately modeled in simulation.

\subsection{Challenging Hand Embodiments}

We investigate the generalization capabilities of \method across different hand embodiments in the main paper. Here, we provide further details on adapting \method to a challenging hand model: the Allegro Hand~\cite{allegrohandurl}, which possesses $K=16$ degrees of freedom. The challenges encountered stem from two primary factors: 1) the Allegro Hand has only four fingers, a significant deviation from the structure of the human hand, and 2) the Allegro Hand is approximately twice the size of a human hand. These morphological discrepancies present substantial challenges in transferring human demonstrations to the Allegro Hand.

\begin{figure}[t!]
    \centering
    \colorbox{white}{\includegraphics[width=\linewidth]{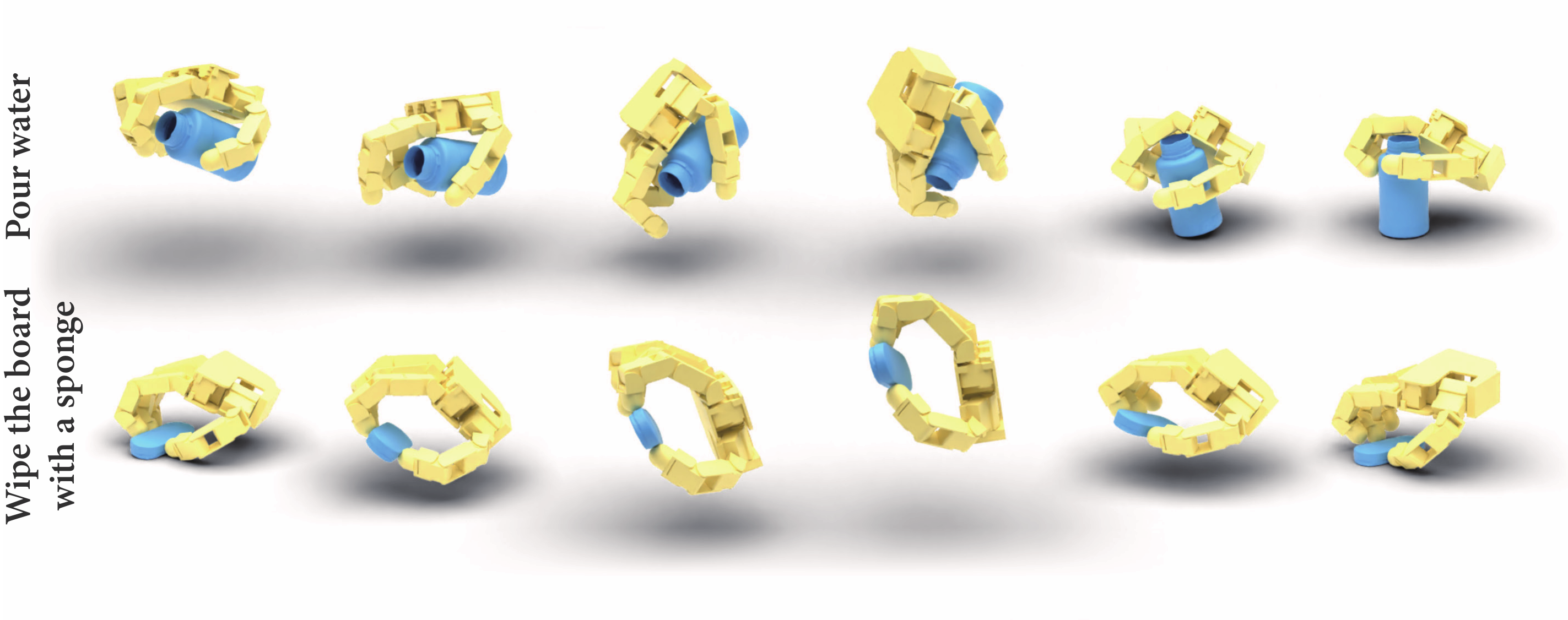}}
    \caption{\textbf{Extending \method to the Allegro Hand.} Despite the Allegro Hand having only four fingers and a significantly larger size, the transferred motion remains stable and natural.}
    \label{fig:supp_allegro}
\end{figure}

To address these challenges, we adaptively modify the fingertip mapping relationships, mapping both the pinky and ring fingers to the same fingertip on the Allegro Hand. Additionally, we relax the fingertip keypoint threshold $\epsilon_{\text{finger}}$ to $8\ cm$ to accommodate the larger dimensions of the Allegro Hand. Successful application of \method to the Allegro Hand is demonstrated in \cref{fig:supp_allegro}.

\subsection{Discussion on the Extension}

To summarize, we present all settings for the extension experiments in \cref{tab:supp_ext}. The green checkmark (\greencheck) indicates the successful transfer of the dataset to the specified hand embodiment, with results included in \dataset. The blue checkmark (\bluecheck) denotes dataset verification, where \method is tested on only a subset of the dataset to assess generalizability. The results demonstrate that our pipeline effectively accommodates various morphological differences across hand embodiments and supports a wide range of tasks, including single-hand manipulation, bimanual articulated object manipulation, and bimanual two-object manipulation.

As discussed in Sec.~{\color{cvprblue} 3.4} of the main paper, FAVOR~\cite{Li_Yang_Lin_Xu_Zhan_Zhao_Zhu_Kang_Wu_Lu_2024} and OakInk-V2~\cite{zhan2024oakink2} represent the largest datasets with the most diverse task types, while the Inspire Hand is distinguished by its high dexterity, stability, cost-effectiveness, and extensive prior use~\cite{jiang2024dexmimicgen, fu2024humanplus, cheng2024open}. Consequently, this setup was chosen for collecting \dataset. However, \method is fully adaptable, and we demonstrate that all of the aforementioned MoCap datasets can be transferred to other robotic hands. We welcome further collaboration from the research community.

\begin{table}[t]
    \rowcolors{6}{WhiteGray}{White}
    \renewcommand{\arraystretch}{1}
    \begin{center}
    \resizebox{1.0\linewidth}{!}
    {
        \setlength{\tabcolsep}{9pt}
        {
        \begin{tabular}{c|cccc}
        \toprule
        \diagbox{Hands}{Datasets} & FAVOR~\cite{Li_Yang_Lin_Xu_Zhan_Zhao_Zhu_Kang_Wu_Lu_2024} & OakInk-V2~\cite{zhan2024oakink2} & GRAB~\cite{taheri2020grab} & ARCTIC~\cite{fan2023arctic}\\
        \midrule
         Inspire~\cite{inspirehandurl}    &\greencheck      & \greencheck         & \bluecheck     &  \bluecheck \\
         Shadow~\cite{shadowhandurl}       &\bluecheck   &\bluecheck        & \bluecheck         & \bluecheck  \\
         Arti-MANO~\cite{MANO:SIGGRAPHASIA:2017}       &\bluecheck    &\bluecheck       & \bluecheck         & \bluecheck  \\
         Allegro~\cite{allegrohandurl}       &\bluecheck    &\bluecheck        & \bluecheck         & \bluecheck  \\
         \bottomrule
        \end{tabular}
        }
    }
    \caption{\textbf{Extensibility of \method.} Arti-MANO refers to the articulated MANO hand used in~\cite{christen2022d}.}
    \label{tab:supp_ext}
    \end{center}
\end{table}

\section{Robustness Evaluation}
\label{sec:supp_robustness}

\begin{table}[b]
    \rowcolors{2}{WhiteGray}{White}
    \renewcommand{\arraystretch}{1}
    \begin{center}
    \resizebox{0.95\linewidth}{!}
    {
        \setlength{\tabcolsep}{9pt}
        {
        \begin{tabular}{cccccc}
        \toprule
        Noise & $E_{r} \downarrow$ & $E_{t} \downarrow$ & $E_{j} \downarrow$ & $E_{ft}\downarrow$ &$SR \uparrow$ \\
        \midrule
         + $\sigma = 0.5\ cm$ & 9.15 & 0.51 & 2.40 &  1.66 & 55.1 / 30.1\\
         + $\sigma = 1.0\ cm$ & 9.56 & 0.57 & 2.87 &  2.13 & 55.3 / 19.5\\
         + $\sigma = 1.5\ cm$ & 9.65 & 0.69 & 3.29 &  2.69 & 46.7 / 39.2\\
         \bottomrule
        \end{tabular}
        }
    }
    \caption{\textbf{Quantitative Results Under Different Noise Levels.} We add the Gaussian noise $\mathcal{N}(0, \sigma^2)$ to the target hand joints poses.}
    \label{tab:meth_noise}
    \end{center}
\end{table}

MoCap data and model-based pose estimation results often contain noise. To assess whether \method can reliably transfer noisy real-world data into stable robotic motions within a simulation environment, we conduct robustness tests. Since \method is designed for general-purpose transfer and does not depend on task-specific reward functions (e.g., the twisting reward proposed in~\cite{lin2024twisting} for the lip-twisting task), noisy object trajectories may introduce instability during the rollout process.
Thus, to evaluate \method's performance under such conditions, we introduce random Gaussian noise into the hand trajectory input and focus on single-hand manipulation tasks. This choice is motivated by the fact that most hand pose estimation methods~\cite{lin2021end,yang2024learning,yang2022artiboost} are optimized for single-hand scenarios.

The results, presented in \cref{tab:meth_noise}, demonstrate that \method maintains acceptable performance even when the noise level reaches up to $1.5\ cm$. These findings highlight the potential of \method for real-world scaling, particularly in applications involving hand pose estimation from web video data, which may implicitly contain a vast array of dexterous manipulation skills.

\section{Time Cost Analysis}
\label{sec:supp_timecost}

\begin{figure}[t!]
    \centering
    \colorbox{white}{\includegraphics[width=0.92\linewidth]{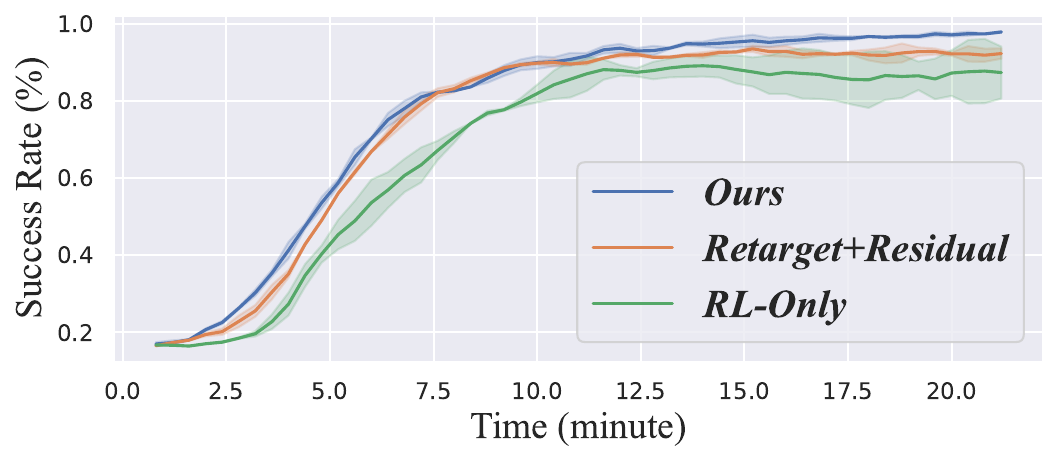}}
    \caption{\textbf{Detailed Efficiency Comparison.} The success rate curves for the ``rotating a mouse" task.}
    \label{fig:supp_cmp_quasisim}
\end{figure}

In Sec.~{\color{cvprblue}4.3} of the main paper, we compare the efficiency of our method with the previous SOTA method, QuasiSim. QuasiSim employs a set of quasi-physical simulations, dividing the transfer process into three primary stages, with each stage requiring approximately 10-20 hours for a 60-frame trajectory \footnote{As reported in the official repository: \url{https://github.com/Meowuu7/QuasiSim}}. Since \method also follows a multi-stage framework, incorporating both a pre-trained hand imitation module and a residual refinement module tailored to physical dynamics, we provide a more comprehensive comparison of efficiency.

For a fair evaluation, we use the official QuasiSim demo data for the ``rotating a mouse" task as a representative example. The success rate curves for three different settings, as discussed in Sec.~{\color{cvprblue}4.3} of the main paper, are shown in \cref{fig:supp_cmp_quasisim}: 1) \textit{RL-Only}: This approach trains the policy network from scratch using RL with our reward design. The curve illustrates the entire training process. 2) \textit{Retarget + Residual} Learning: Inspired by~\cite{zhaodexh2r}, this method retargets human hand poses to initial dexterous hand poses via keypoint alignment~\cite{qin2023anyteleop}, followed by residual learning for refinement. The retargeting process is performed via parallel optimization and only requires approximately several minutes on a single GPU to optimize full sequence. The training curve for the residual learning stage is represented by the orange line. 3) \method: We pre-train the hand imitation model on a large-scale training dataset, as described in Sec.~{\color{cvprblue}3.2}, which takes approximately 1.5 days on a single GPU to obtain the reusable imitator. The residual learning stage training curve is shown by the blue line.

From the results in \cref{fig:supp_cmp_quasisim}, we observe that for the relatively simple task of ``rotating a mouse", the \textit{Retarget + Residual} method achieves performance comparable to \method but requires slightly more time to converge. The \textit{RL-Only} approach, while yielding suboptimal performance compared to the other methods, still produces acceptable motions within 20 minutes. This indicates that our reward design effectively accelerates the training process, facilitating faster convergence.

\section{Details of \method Settings}
\label{sec:supp_details_method}

\subsection{Correspondence Between Human Hand and Dexterous Hand}

Due to the significant morphological differences between human hands and dexterous robotic hands, we manually establish correspondences between them.
For the human hand's fingertip keypoints, we select the midpoint of the three tip anchors as defined in~\cite{yang2024learning}. For the dexterous hands, given their varying shapes, we define the fingertip keypoints as the points of maximum curvature along the central axis of the finger pads, as these points are most likely to contact objects. For other keypoints, such as the wrist and phalanges, we intuitively align the rotation axes of the human joints with those of the robotic joints. For further details, please refer to our code implementation.

In addition, regarding the articulated MANO model, the original human hand model MANO~\cite{MANO:SIGGRAPHASIA:2017} has 45-DoF, which presents extreme challenges for RL-based policies due to the vast exploration space. To mitigate this, we follow the approach in~\cite{yang2024learning} by constraining certain DoFs and fixing the hand collision meshes, thereby reducing the original MANO model to a 22-DoF articulated MANO.

\subsection{Details of Training Parameters}
\begin{table}[b]
    \rowcolors{6}{WhiteGray}{White}
    \renewcommand{\arraystretch}{1}
    \begin{center}
    \resizebox{0.7\linewidth}{!}
    {
        \setlength{\tabcolsep}{9pt}
        {
        \begin{tabular}{c|c|c}
        \toprule
        Figners & weight $w_f$ & decay rate $\lambda_f$\\
        \midrule
         Thumb    & 0.5, 0.3, 0.3, 0.9      & 50, 40, 40, 100        \\
         Index    & 0.5, 0.3,0.3, 0.8      & 50, 40, 40, 90       \\
            Middle   & 0.5, 0.3,0.3, 0.75      & 50, 40, 40, 80        \\
            Ring     & 0.5, 0.3,0.3, 0.6      & 50, 40, 40, 60        \\
            Pinky   & 0.5, 0.3,0.3, 0.6      & 50,40, 40, 60        \\
        
         \bottomrule
        \end{tabular}
        }
    }
    \caption{\textbf{Hyperparameters for the Finger Reward.} The weight $w_f$ and decay rate $\lambda_f$ are used to balance the importance of each finger. Each cell in the table contains four values, representing the parameters for the proximal, intermediate, distal, and tip joints, respectively. For anatomical definitions, please refer to~\cite{yang2024learning}.}
    \label{tab:supp_param}
    \end{center}
\end{table}

In this section, we present the core parameters of our reward functions in \method. The reward parameters for $r^t_{\text{finger}}$ in Eq.~({\color{cvprblue}1}) of the main paper are summarized in \cref{tab:supp_param}. These parameters are determined based on the observation that the thumb, index, and middle fingers play a pivotal role in grasping and manipulation tasks, as they statistically interact with objects more frequently than other fingers~\cite{zhan2024oakink2, Brahmbhatt_2020_ECCV, Brahmbhatt_2019_CVPR}. Consequently, the weights are assigned according to the contact frequency.
In our implementation, if a dexterous hand lacks a specific finger or joint (\eg, the Inspire Hand does not have distal joints), the corresponding parameters are set to zero.
For the contact reward $r^t_{\text{contact}}$ in Eq.~({\color{cvprblue}2}) of the main paper, we set both parameters, $w_c$ and $\lambda_c$, to 1.\del{ The full code and dataset will be released for further implementation details.}

\subsection{Details of Simulation Parameters}

In the Isaac Gym environment, configuring physical properties significantly influences the success rate of transfer. Alongside domain randomization (DR) during training, we set physical constants as follows. For certain objects in OakInk-V2 \cite{zhan2024oakink2}, we obtained actual masses by directly measuring them in collaboration with the dataset authors. For the remaining objects, we assigned a constant density of $200\ kg/m^3$, approximating the average density of low-fill-rate 3D-printed models. Using this density, we recalculated the objects' masses and moments of inertia.

It is worth noting that human skin is elastic. When grasping objects, fingertip skin undergoes slight deformations, enhancing contact with object surfaces and generating suitable friction, whereas dexterous robotic hands lack this behavior. Previous kinematics-based grasp generation methods \cite{jiang2021hand,li2024semgrasp} often permit slight penetration between fingertips and object surfaces to improve interaction stability (for detailed discussion, please refer to \cite{jiang2021hand}). Therefore, to compensate for the absence of skin deformation in simulation, we set the friction coefficient $\mathcal{F}$ slightly higher than the real-world value. Accurately simulating contact-rich scenarios remains an area for future exploration.

\begin{table}[b]
    \rowcolors{6}{WhiteGray}{White}
    \renewcommand{\arraystretch}{1}
    \begin{center}
        \label{table:supp_prompt_mllm}
        \resizebox{1.0\linewidth}{!}{
            \begin{tcolorbox} 
                \scriptsize
                \texttt{\underline{assemble}, \underline{brush whiteboard}, \underline{cap}, \underline{cap the pen}, \underline{close book}, \underline{close gate}, \underline{close laptop lid}, \underline{cut}, \underline{flip close tooth paste cap}, \underline{flip open tooth paste cap}, heat beaker, heat test tube, \underline{hold}, \underline{hold test tube}, \underline{ignite alcohol lamp}, \underline{insert lightbulb}, \underline{insert pencil}, \underline{insert usb}, \underline{open gate}, \underline{open laptop lid}, \underline{place asbestos mesh}, \underline{place inside}, place on test tube rack, place onto, \underline{place test tube on rack with holder}, \underline{plug in power plug}, \underline{pour}, \underline{pour in lab}, press button, \underline{put flower into vase}, \underline{put off alcohol lamp}, \underline{put on lid}, \underline{rearrange}, remove from test tube rack, \underline{remove lid}, \underline{remove pencil}, \underline{remove power plug}, \underline{remove test tube}, remove test tube from rack with holder, \underline{remove the pen cap}, \underline{remove usb}, \underline{scoop}, \underline{scrape}, \underline{screw}, shake lab container, \underline{sharpen pencil}, \underline{shear paper}, \underline{spread}, \underline{squeeze tooth paste}, \underline{stir}, \underline{stir experiment substances}, \underline{swap}, \underline{take outside}, trigger lever, \underline{uncap}, \underline{uncap alcohol lamp}, \underline{unscrew}, \underline{use mouse}, \underline{wipe}, write on paper, \underline{write on whiteboard}}
            \end{tcolorbox}
    }
    \caption{List of tasks in the \dataset dataset. Tasks with \underline{underlined} names usually require bimanual manipulation.}
    \label{tab:supp_tasks}
    \end{center}
\end{table}

\section{\dataset Statistics}
\label{sec:supp_dataset_stats}

To the best of our knowledge, no prior work has collected a large-scale bimanual manipulation dataset in which all trajectories are directly transferred from real human demonstrations without the use of teleportation. Leveraging the efficiency and precision of \method, our dataset, \dataset, comprises 3.3K diverse manipulation trajectories across 61 distinct tasks, as detailed in \cref{tab:supp_tasks}. To ensure stability during simulation, we fix the object meshes to a watertight state using ManifoldPlus~\cite{huang2020manifoldplus} and may slightly adjust the object size to enhance object-object interactions (e.g., the cap and body of the bottle).

Additionally, we provide sample data on our website, showcasing trajectories generated from our policy in simulation. A simple first-order low-pass filter ($\alpha=0.4$) is applied to the rollouts, effectively reducing jitter with minimal impact on tracking accuracy.

\section{Details of Rearrangement Policy Learning}
\label{sec:supp_imitation}

\begin{figure}[t]
    \centering
    \includegraphics[width=1.0\linewidth]{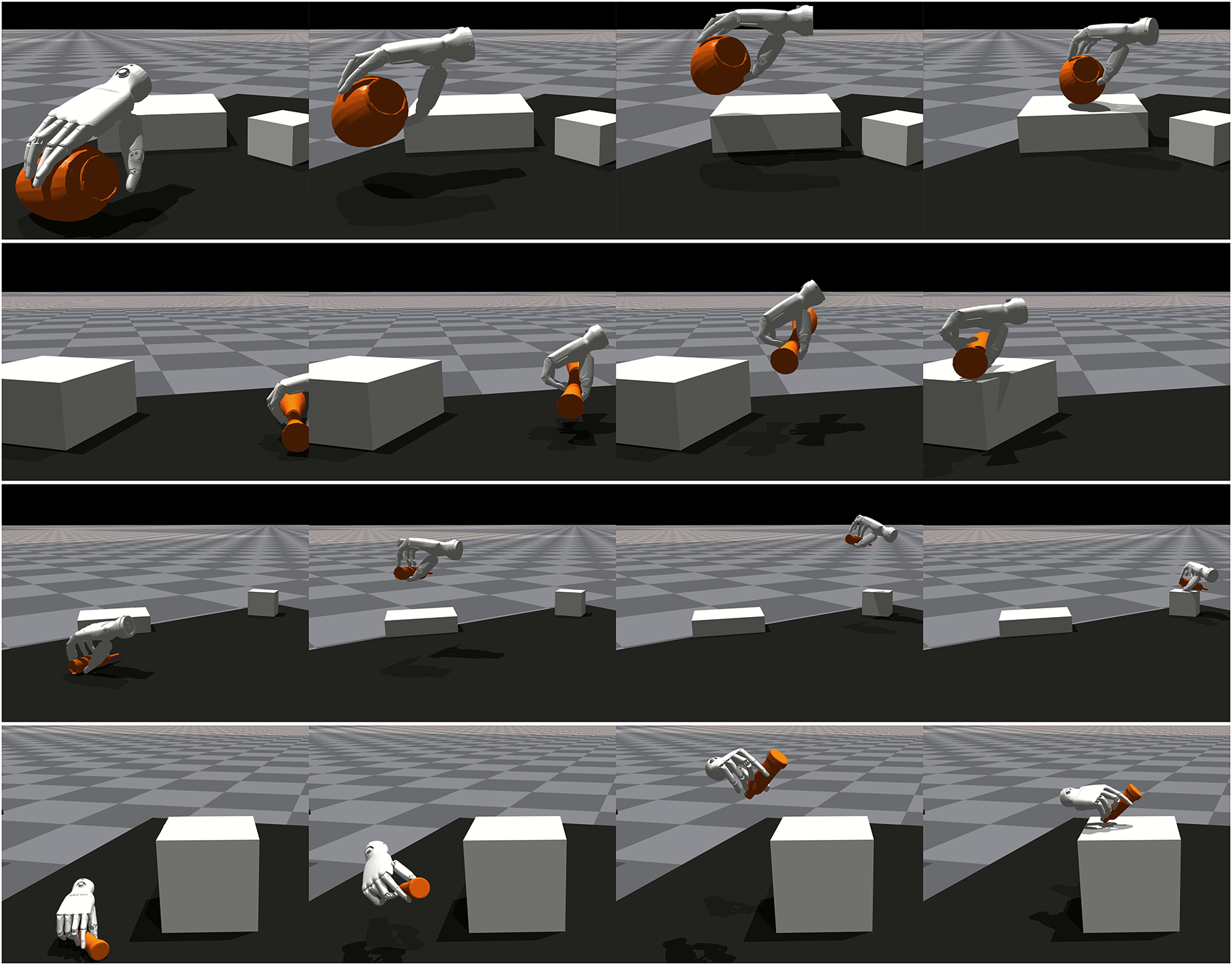}
    \caption{\textbf{Qualitative Results of Rearrangement Policy Learning.} The policy successfully moves the bottle to the goal position. Results are directly visualized in the IsaacGym environment, highlighting distinctions between these policies and \method's rollouts.}
    \label{fig:supp_imitation-example}
\end{figure}

As discussed in Sec.~{\color{cvprblue}4.7} of the main paper, we benchmark \dataset using four data-driven imitation learning methods on the \textit{moving a bottle to a goal position} task.

The primary challenge in this task is to enable the dexterous hand to maintain a stable grasp on the object while smoothly placing it at the specified goal position. We evaluate the dataset using four methods: IBC~\cite{florence2022implicit}, BET~\cite{shafiullah2022behavior}, and Diffusion Policy~\cite{chi2023diffusion}, which include both UNet- and Transformer-based architectures. These policies are trained for 500 epochs using the Adam optimizer with a learning rate of $1 \times 10^{-4}$, while all other hyperparameters remain at their default settings.

The dimensions of the observation and action spaces for these policies are provided in \cref{tab:supp_imitation-space}. The observation space includes the current object state $\{\mbold{p}_{\cobj}, \dot{\mbold{p}}_{\cobj}\}$, the hand wrist state $\{\mbold{w}_{\mbold{d}}, \dot{\mbold{w}}_{\mbold{d}}\}$, hand joint angles $\mbold{q}_{\dexhand}$, and the goal poses for both the object $\mbold{g}_{\cobj}$ and the hand wrist $\mbold{g}_{\mbold{w}}$. The action $\mbold{a} = \{\mbold{a}_{\mbold{q}}, \mbold{a}_{\mbold{w}}\} \in \mbold{\mathcal{A}}$ specifies the target hand joint angles and wrist poses using a PD controller. Note that PD control is used for wrist poses rather than a 6-DoF force, as is done in \method.

We evaluate the policies' performance on previously unseen goal positions within the IsaacGym environment~\cite{makoviychuk2021isaac}. A rollout is considered successful if the object's distance from the goal position is within $10\ cm$; otherwise, it is classified as a failure. Qualitative results are presented in \cref{fig:supp_imitation-example}, while quantitative results are summarized in Tab.~{\color{cvprblue}2} of the main paper.

\begin{table}[ht!]
    \centering
    \small
    \begin{subtable}{\linewidth}
        \centering
        \begin{tabular}{cc}
            \toprule
                \textbf{Observation} & \textbf{Dimensions} \\
            \midrule
                Hand joint angles $\mbold{q}$ & 12 \\
                Hand wrist state $\{\mbold{w}_{\mbold{d}}, \dot{\mbold{w}}_{\mbold{d}}\}$ & 13 \\
                Object state $\{\mbold{p}_{\cobj}, \dot{\mbold{p}}_{\cobj}\}$ & 13 \\
                Object pose goal $\mbold{g}_{\cobj}$ & 7 \\
                Hand wrist pose goal $\mbold{g}_{\mbold{w}}$ & 7 \\
            \bottomrule
        \end{tabular}
        \caption{Observation space.}
    \end{subtable}%
    \\%
    \begin{subtable}{\linewidth}
        \centering
        \vspace{6pt}
        \begin{tabular}{cc}
            \toprule
                \textbf{Action} & \textbf{Dimensions} \\
            \midrule
                Hand joint angles $\mbold{a}_{\mbold{q}}$  & 12 \\
                Hand wrist pose $\mbold{a}_{\mbold{w}}$ & 7 \\
            \bottomrule
        \end{tabular}
        \caption{Action space.}
    \end{subtable}

    \caption{\textbf{Observation and Action Definitions for the Imitation Policy.} The policy's 7-dimensional pose includes both position and orientation, represented as XYZW quaternions. The policy's 13-dimensional state extends this pose by incorporating both linear and angular velocities.}
    \label{tab:supp_imitation-space}
\end{table}
\end{appendix}

\end{document}